\documentclass{article}

\usepackage{float}
\usepackage{wrapfig}          

\usepackage{amsmath}
\usepackage{amssymb}
\usepackage{mathtools}
\usepackage{graphicx}
\usepackage{amsthm}

\usepackage{xcolor}
\usepackage{float}

\usepackage{subcaption}

\usepackage[dvipsnames]{xcolor}
\newcommand\NavyBlue{\color{NavyBlue}}
\newcommand\Orange{\color{Orange}}
\usepackage{microtype}
\usepackage{graphicx}
\usepackage{subcaption}
\usepackage{booktabs} 
\usepackage[ruled,vlined,linesnumbered]{algorithm2e}
\usepackage{multirow}
\newcommand{\ours}{StFT\xspace}

\usepackage{hyperref}

\usepackage[preprint]{neurips_2025}

\usepackage[capitalize,noabbrev]{cleveref}

\theoremstyle{plain}

\theoremstyle{definition}

\theoremstyle{remark}

\usepackage[textsize=tiny]{todonotes}

\title{StFT: Spatio-temporal Fourier Transformer for Long-term Dynamics Prediction}

\author{%
  Da Long \\
  School of Computing\\
  University of Utah\\
  \And
  Shandian Zhe \\
  School of Computing\\
  University of Utah\\
  \AND
  Samuel Williams \\
  AMCR Division \\
  Lawrence Berkeley National Laboratory\\
  \And
  Leonid Oliker \\
  AMCR Division \\
  Lawrence Berkeley National Laboratory\\
  \And
  Zhe Bai
  $^{*}$\\
  Applied Mathematics and Computational Research Division \\
  Lawrence Berkeley National Laboratory\\
  Berkeley, CA 94720, USA
}

\begin{document}
\maketitle

\begin{abstract}
Simulating the long-term dynamics of multi-scale and multi-physics systems poses a significant challenge in understanding complex phenomena across science and engineering. The complexity arises from the intricate interactions between scales and the interplay of diverse physical processes, which manifest in PDEs through coupled, nonlinear terms that govern the evolution of multiple physical fields across scales. Neural operators have shown potential in short-term prediction of such complex spatio-temporal dynamics; however, achieving stable high-fidelity predictions and providing robust uncertainty quantification over extended time horizons remains an open and unsolved area of research. These limitations often lead to stability degradation with rapid error accumulation, particularly in long-term forecasting of systems characterized by multi-scale behaviors involving dynamics of different orders. To address these challenges, we propose an autoregressive Spatio-temporal Fourier Transformer (\ours), in which each transformer block is designed to learn the system dynamics at a distinct scale through a dual-path architecture that integrates frequency-domain and spatio-temporal representations. By leveraging a structured hierarchy of \ours blocks, the resulting model explicitly captures the underlying dynamics across both macro- and micro- spatial scales. Furthermore, a generative residual correction mechanism is introduced to learn a probabilistic refinement temporally while simultaneously quantifying prediction uncertainties, enhancing both the accuracy and reliability of long-term probabilistic forecasting. Evaluations conducted on three benchmark datasets (plasma, fluid, and atmospheric dynamics) demonstrate the advantages of our approach over state-of-the-art ML methods. 

\let\thefootnote\relax
\footnotetext{\hspace{-5pt}$^*$ Corresponding author (zhebai@lbl.gov)}
\fontsize{10}{12}\selectfont

\end{abstract}


\section{Introduction}

Predicting the long-term spatio-temporal dynamics of systems governed by partial differential equations (PDEs) is a cornerstone of scientific and engineering research, with broad applications in fields such as earth system modeling, plasma science, fluid dynamics, and beyond. Traditional approaches rely heavily on numerical solvers, which discretize the domain and iteratively solve PDEs using methods including finite element, finite volume and spectral methods~\citep{tadmor2012review}. While effective in many scenarios, these techniques face limitations when applied to multiphysics systems characterized by complex dynamics and multiscale behaviors. They require substantial computational resources and exhibit poor scalability with increasing problem size, rendering them impractical for high-dimensional, large-scale, or long-term physics systems due to excessive computational costs and memory demands.

Recent advances in deep learning have revolutionized the field of PDE modeling by introducing data-driven methodologies that significantly accelerate computations for science while maintaining high accuracy. Inspired by the universal approximation theorem~\citep{chen1995universal}, neural operators that learn the mapping between two function spaces have demonstrated great success in simulating various PDE systems across multiple scientific disciplines without retraining for new conditions~\citep{li2020fourier,lu2021learning}. Building on the success of transformers in natural language processing and computer vision~\citep{vaswani2017attention,dosovitskiy2020image}, transformer-based neural operators process multiple input functions while enabling arbitrary querying of output function locations, offering enhanced flexibility in handling complex functional mappings~\citep{hao2023gnot,li2022transformer}. 
A series of neural operators have been developed to address complex scientific problems, including weather forecasting, turbulent fluid dynamics, and boiling phenomena~\citep{pathak2022fourcastnet,li2023long,bi2023accurate,hassan2023bubbleml,lin2021operator}. 

Despite the success of these methods, accurate and long-term predictions of complex physical systems remain challenging, primarily due to the requirements for numerical stability, high-fidelity modeling, and reliable uncertainty quantification over extended horizons. The inherent multi-scale nature and multi-physics complexity of such systems necessitate methodologies that can efficiently represent and integrate dynamics across disparate spatial and temporal scales while simultaneously capturing the complex interactions between distinct physical processes, such as the influence of micro-scale turbulence on macro-scale flow in fluids and combustion~\citep{peters2009multiscale,natrajan2007statistical}. For large-scale atmospheric pressure systems, high-pressure ridges and low-pressure troughs play a crucial role in shaping local weather patterns; inaccurate representation of those structures can cause significant errors in forecasting rainfall, wind speed, and temperature~\citep{wang2006large,barlow2019north}. In magnetically confined plasmas, multiphysics arises from the coupling of physical processes that govern plasma behavior, including electromagnetic fields, turbulence transport, thermodynamics, and particle interactions that are potentially coupled with kinetic models in high-fidelity simulations. The magnetohydrodynamic (MHD) instabilities caused by current or pressure gradients can limit burning plasma performance, and threaten fusion device integrity~\citep{von1974studies,graves2012control,seo2024avoiding}.
Furthermore, integrating uncertainty quantification (UQ) into modeling frameworks is essential for assessing the confidence and reliability of predictions in such complex systems~\citep{cheung2011bayesian,scher2018predicting,kruger2024thinking}. 
Although neural operators present advantages over traditional approaches, they still encounter challenges associated with the demands for scientific fidelity and stability, especially when the underlying physics exhibit rapid changes or high-frequency components. These issues are further intensified in high-resolution simulations of multi-scale scenarios. Recent efforts to address these limitations include $P^{2}C^{2}$Net, which encodes a high-order numerical scheme with boundary condition encoding into neural networks~\citep{NEURIPS2024_7f605d59}, and Dyffusion, which trains a forecasting network and an interpolation network that allows for continuous time sampling and multi-step prediction for long-range forecasting~\citep{ruhling2023dyffusion}.
However, most existing neural operators lack built-in mechanisms for uncertainty quantification, which is particularly critical for reliable modeling of long-term dynamics, where even small errors can propagate across scales and result in significant inaccuracies. 


Existing approaches for predicting spatio-temporal dynamics can be broadly be classified into two primary categories. The first category comprises models that directly forecast future states at fixed time horizons using  a sequence of past observations~\citep{wang2024bridging,kontolati2024learning}. The second category includes models that utilize an autoregressive manner, which addresses challenges of scaling and fitting complexities as a continuous-time emulator~\citep{pathak2022fourcastnet,ruhling2023dyffusion,lippe2023pde,mccabe2023towards}. Generally, prediction errors incurred in the short term can accumulate, leading to instability and reduced accuracy in long-term forecasts. To mitigate these issues, previous work has proposed techniques such as the pushforward trick, invariance preservation, and iterative refinement~\citep{mccabe2023towards,lippe2023pde,brandstetter2022message}. Nevertheless, the development of multi-scale modeling frameworks for long-term dynamic prediction remains crucial for capturing the interactions across scales and enhancing prediction accuracy. Concurrently, incorporating uncertainty quantification is critical for identifying the spatial and temporal regions where predictive confidence deteriorates, especially in complex systems where localized uncertainties can influence global dynamics over time.

In this work, we introduce an autoregressive \textbf{Spatio-temporal Fourier Transformer (StFT)}, for long-range forecasting of multi-scale and multi-physics systems. 
At each level of the spatial hierarchy, one StFT block models the physical dynamics associated with a distinct spatial scale or receptive field, as inferred from the spatiotemporal data. 
Each StFT block adopts a dual-path architecture: (1) \textbf{the frequency path} captures large-scale dynamics by operating in the Fourier domain, focusing on low-frequency components that are critical for modeling global behavior. (2) \textbf{the spatio-temporal path} operates in the full physical space incorporating all spatio-temporal features to capture fine-scale features.
Through a hierarchical composition of StFT blocks across multiple scales, augmented by a generative residual correction block, the resulting model learns the intricate interactions both within the same scales and across different scales. Moreover, it produces uncertainty estimates at each spatial and temporal point, enabling assessment of prediction confidence throughout the forecast. The cascading StFT blocks enable our model to predict high-resolution dynamics across a spectrum of varying scales in correlated physical processes. 
By integrating StFT within an auto-regressive framework, our method achieves superior accuracy in long-term predictions compared to existing state-of-the-art autoregressive baselines.
Our contributions are summarized as follows:
\begin{itemize}
    \item We propose Spatio-temporal Fourier transformer (StFT), a novel ML model that learns underlying dynamics across spatial scales for multi-physics systems via a dual-path (frequency and spatio-temporal path) architecture, which effectively captures both the global, large-scale structures and local, fine-scale features.
    \item We propose StFT-F, which incorporates a probabilistic residual correction mechanism to refine the forecasting of StFT temporally and provide pointwise uncertainty quantification.
    \item We propose an overlapping tokenizer and a detokenizer that share regions between adjacent patches, improving spatial smoothing and reducing discontinuity artifacts.
    \item We demonstrate the effectiveness of StFT in an autoregressive framework on a diverse set of applications including the plasma, fluid, and atmospheric dynamics. Evaluating performance across variables, StFT outperforms the best baselines. Its probabilistic variant, StFT-F further improves average forecasting accuracy by 5\% and produces uncertainty estimates that are empirically calibrated, as demonstrated through confidence-based evaluation.
    
\end{itemize}

\section{Related Work}
\textbf{Neural Operators.} Neural operator architectures and their variants have been proposed, including Fourier neural operators~\citep{li2020fourier,gupta2021multiwavelet,tran2021factorized,cao2023lno,li2023fourier,rahman2022u}, DeepONet~\citep{lu2021learning,wang2021learning,jin2022mionet,wang2022improved,kontolati2024learning,prasthofer2022variable}, transformer based operators~\citep{hao2023gnot,cao2021choose,li2022transformer}, and image-to-image operators~\citep{gupta2022towards,long2024arbitrarily}. U-Net, a fundamentally hierarchical structure model, which has inspired several neural operators ~\citep{rahman2022u,liu2022ht,gupta2022towards}, allows solutions to multi-scale PDEs by hierarchically aggregating feature representations of progressively coarser spatial resolutions. Recent work in computer vision~\citep{liu2021swin,fan2021multiscale,zhang2022nested} have introduced methods for extracting multi-scale features through hierarchical architectures. However, these hierarchical models do not explicitly model or forecast the multi-scale structures of physical processes, which limit the transparency and interpretability of their representations across scales. In contrast, our method begins with a coarse approximation that captures large-scale, low-frequency phenomena, and incrementally refines the representation over layers to resolve finer details. These structured decompositions enable error diagnosis, enhance interpretability of model performance of different scales, and allow for targeted improvements with an explicit refinement mechanism.

\textbf{Generative Models.} Generative models, especially diffusion models have demonstrated success in various domains, including vision, audio, robotics~\citep{ho2020denoising,song2020score,tian2024transfusion,kong2020diffwave,wolleb2022diffusion}, and relevance to spatio-temporal dynamics prediction~\citep{ho2022video,voleti2022mcvd,singer2022make,ruhling2023dyffusion}. As an alternative approach in generative modeling, flow matching has been introduced to support efficient sampling and has since been applied to video generation~\citep{lipman2022flow,liu2022flow,albergo2022building,polyak2024movie,esser2403scaling}. While video generation typically explores a range of creative and diverse possibilities from text or image prompts, forecasting spatial-temporal dynamics driven by PDEs necessitates more than mere statistical resemblance - it requires each prediction is firmly grounded in the underlying physics. To achieve accurate forecasting while capturing the inherent stochasticity of physical processes, our work incorporates a flow matching block following the proceeding \ours blocks. This enables the model to align its prediction distribution with the underlying physical dynamics and generate calibrated uncertainty estimates via confidence-based metrics.

\section{Method}

\begin{figure}[t]
	\centering
    \includegraphics[width=.95\linewidth]{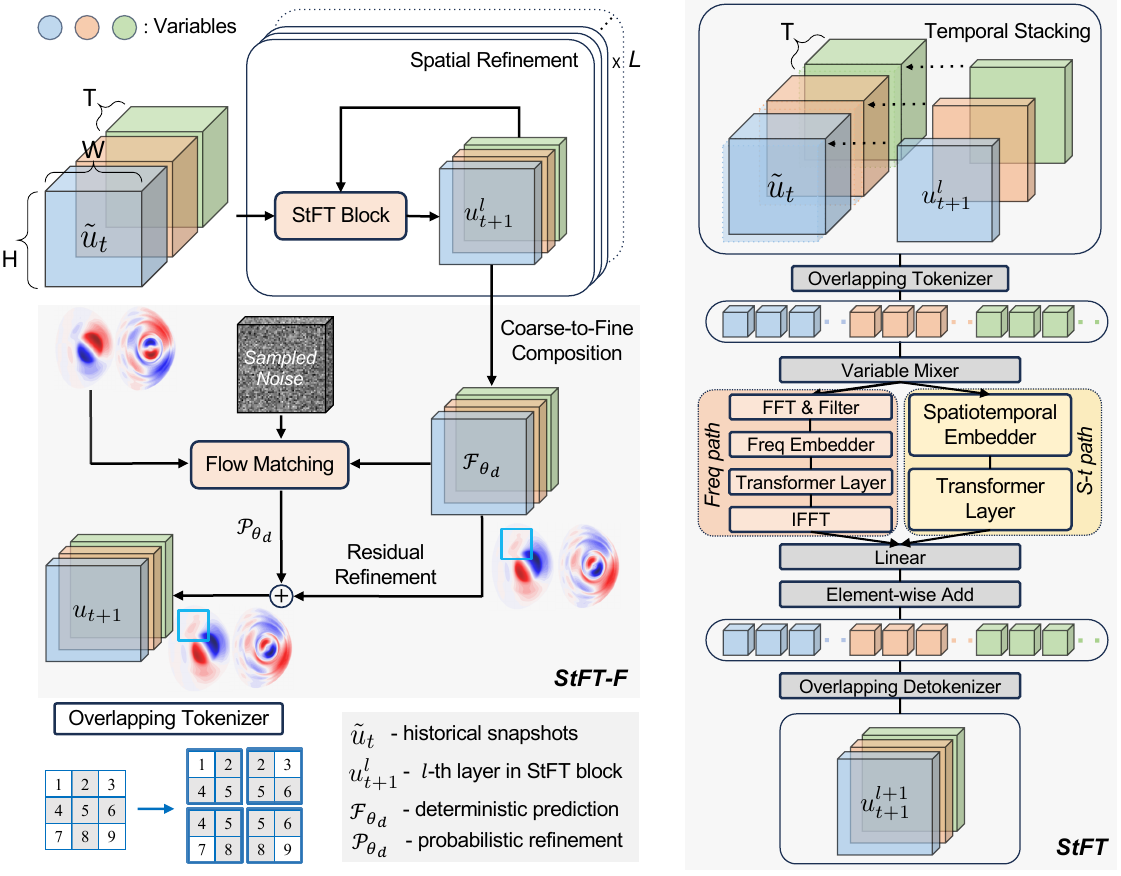}
	\caption{\small Left: 
\textbf{Overview of the proposed StFT and StFT-F model}. The model predicts $u_{t+1}$ using the past $k$ snapshots $\tilde{u}_{t}$, employing $L$ spatial refinements from coarse to fine scales through the proposed Spatiotemporal Fourier Transformer (StFT) blocks. Bottom left: an illustration of the overlapping tokenizer, where the patch size is $2 \times 2$, and the overlapping number is $1 \times 1$. Right: \textbf{Illustration of the proposed \ours block}. First, the past snapshots $\tilde{u}_t$ and a coarser prediction from the previous layer are temporally stacked. The stacked discretizations are passed to the overlapping tokenizer to generate tokens for each variable. Next, tokens corresponding to different variables in the same spatial are mixed through a variable mixer. Two paths of transformation, frequency path and spatiotemporal path, process the frequency embeddings and the spatiotemporal embeddings respectively. The finer prediction for timestamp $t+1$ is obtained after passing through the overlapping detokenizer.}
\label{fig:stft}  
\end{figure}

\textbf{Formulation. } We consider an autoregressive formulation for long-term multi-scale spatiotemporal physical processes. We define a vector $\tilde{u}_t$ representing the historical snapshots of the multi-physics variables at timestamps from $t-k+1$ to $t$, in a total of $k$ snapshots of $\tilde{u}_t = [u_t, u_{t-1}, \dots, u_{t-k+1}]$ specifically. We formulate the probabilistic one-step forward neural operator StFT-F as
\begin{align}
        u_{t+1} = \mathcal{F}_{\theta_d}(\tilde{u}_{t}) + r_{t+1},
        r_{t+1} \sim \mathcal{P}_{\theta_g}(r | \tilde{u}_{t},\mathcal{F}_{\theta_d}(\tilde{u}_{t})),
        \label{eq:onestep}
\end{align} 
where $\mathcal{F}_{\theta_d}$ denote the \ours operator, a deterministic forecasting parameterized by $\theta_d$, and $\mathcal{P}_{\theta_g}$ is the generative flow matching block parameterized by $\theta_g$ for refining the forecasting of \ours while quantifying uncertainty estimates. $\mathcal{F}_{\theta_d}$ represents the deterministic evolution of the system that encapsulates the dynamics from multi-scale spatial refinement in StFT. The residual refinement $\mathcal{P}_{\theta_g}(r | \tilde{u}_{t},\mathcal{F}_{\theta_d}(\tilde{u}_{t}))$ captures the probabilistic nature of the residual from the generative model. It represents a prediction distribution conditioned on the current state $\tilde{u}_{t}$ and the deterministic prediction $\mathcal{F}_{\theta_d}(\tilde{u}_{t})$, modeling the uncertainty or variations that missed by the deterministic component. The residual $r_{t+1}$ calculates deviations from the deterministic prediction, and its distribution allows the model to account for noise or inherent stochasticity in the physical processes. Therefore, by sampling residual $r_t$, our model learns stochastic trajectories from data. Besides providing prediction uncertainties, these stochastic trajectories can help study the long-term behavior, stabilities, and bifurcations in stochastic systems ~\citep{lucor2003predictability,kramer2022probabilistic}.

Figure~\ref{fig:stft} presents the overview of the StFT-F, the overlapping tokenizer, and the design of the \ours block. Algorithm ~\ref{alg:overview} and ~\ref{alg:stft_details} detail the design of our model. The following subsections introduce StFT with the overlapping tokenizer/detokenizer and the residual correction mechanism based on flow matching. 

\subsection{\ours Block}



\textbf{Overlapping tokenizer.} Each \ours block first tokenizes the discretized functions $\tilde{u}_{t}$ of shape $T\times W\times H\times C$, where $T$ is the temporal dimension, $W, H$ are the spatial dimensions, and $C$ denotes the number of physical variables. We apply tokenization along the spatial dimensions for each variable. To enhance spatial continuity while minimizing visual artifacts, we propose to use an overlapping tokenizer (OLT) and detokenizer (OLDT) that allows adjacent patches to share boundaries through overlapping regions. For instance, as shown in Figure~\ref{fig:stft}, a $3\times3$ input generates four $2\times2$ patches with a $1\times1$ overlap, where the overlapping areas (indicated in gray) are shared between patches. During detokenization, overlapping regions are reconstructed by averaging the corresponding values from neighboring patches. This strategy effectively mitigates discontinuity issues at the patch boundaries, which is particularly important for accurately representing smooth and continuous target functions. Furthermore, incorporating shared boundaries into patch embeddings enriches feature representation and extraction from fine-scale structures.

\textbf{Variable mixer.} Each \ours block is designed to handle a specific scale; therefore, by employing a specific patch size, we partition the input at a corresponding level of granularity. To ensure that the first block captures the coarsest features or the largest scale, we set the patch size $p_w^1 \times p_h^1$ to a large value, allowing it to model broad spatial structures effectively. As a result,  $\mathcal{O}(\frac{W}{p_w^1}\times \frac{H}{P_h^1}\times C)$ patches are fed into the variable mixer, where patches corresponding to different physical processes but sharing the same spatial domain are mixed into a single token. Following this step, two transformation paths are performed: one is in the spatio-temporal domain, which operates self-attention on spatio-temporal embeddings; the other is in the frequency domain, which operates on frequency embeddings.

\textbf{Frequency embeddings.}
The tokens are first processed by a 2D/3D Fourier transform, where only low-frequency components are retained. These low-frequency components are then passed through a frequency embedder to obtain frequency embeddings $f_{t}$. Subsequently, these frequency embeddings are fed to the standard transformer layers for mixing information and nonlinear transformation in the frequency domain. Finally, an inverse 2D/3D Fourier transform and a linear projection are applied to map the frequency embeddings back to the spatio-temporal domain.

\textbf{Spatio-temporal embeddings.}
The same set of tokens first pass through a spatio-temporal embedder, after which the spatio-temporal embeddings $e_{t}$ are processed by multiple standard transformer layers for mixing correlations and nonlinear transformation. Finally a separate linear projection is applied to get predictions for each patch of $u_{t+1}$.

Next, an overlapping detokenizer yields the first block prediction $u_{t+1}^1$. Each token represents a significant portion of the historical snapshots, encapsulating macroscopic structural features. This coarse-level partitioning reduces the complexity of modeling fine-grained details. By maintaining a lower granularity, the model prioritizes structural coherence over extraneous details, enabling it to focus on capturing and predicting global relationships between regions more effectively. 

\subsection{A Hierarchy of \ours Blocks}
\begin{wrapfigure}{r}{0.28\textwidth}
  \begin{center}
\vspace{-45pt}
\includegraphics[width=.98\linewidth]{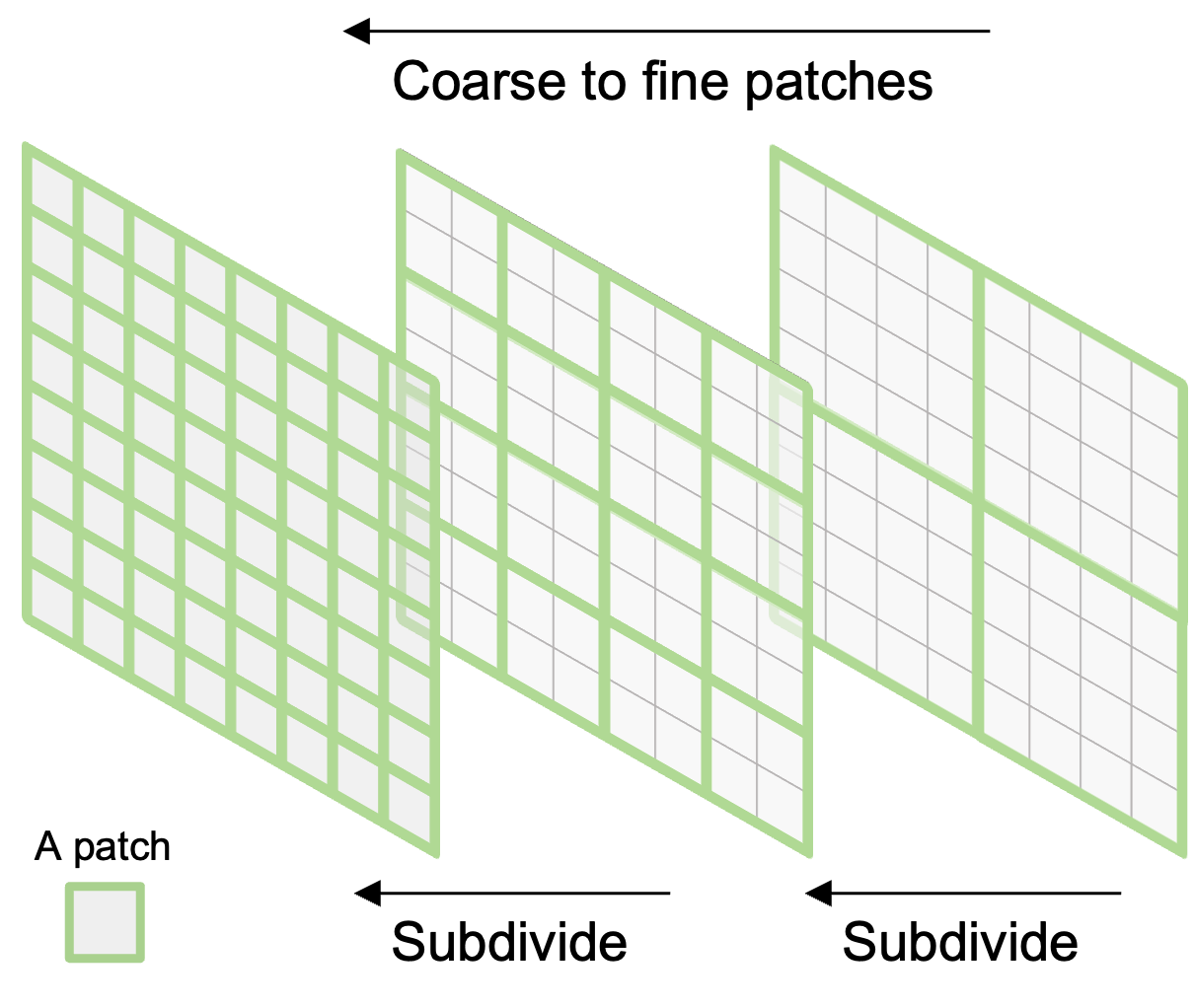}
  \end{center}
  \caption{\small A patch is subdivided into smaller patches for a hierarchical learning.} 
\vspace{-40pt}
\end{wrapfigure}
In the subsequent \ours blocks, we shift our focus to smaller scales with fine details. Consequently, we concatenate the prediction $u_{t+1}^1$ with the input $\tilde{u}_{t}$, and consider this combination as the input for the next \ours block. We further subdivide each patch from the previous \ours block into smaller patches. 
The smaller patches allow for less information to be aggregated within a single patch, thereby minimizing the risk of losing local variations and enhancing the richness and informativeness of the fine-scale representation. By leveraging the finer granularity of the patches to focus on smaller regions, it allows the model to better localize features and capture their details. In addition, conditioning on the coarser prediction allows the models to iteratively refine its estimates, beginning with a broad global summary of the corresponding regions.
Through the repeated subdivision of the patches, the model progressively refines its predictions across multiple scales. As shown in Figure~\ref{fig:stft}, the StFT model predicts $u_{t+1}$ and applies a total of $L$ multi-scale spatial refinements.

\subsection{Residual Refinement and Uncertainty Estimation Based on Flow Matching} 
Finally, the model refines its deterministic predictions through a rectified flow block, which belongs to the family of flow matching models~\citep{liu2022rectified,liu2022flow}. Flow matching is formulated as an ordinary differential equation in time $\tau\in[0,1]$, $\frac{d}{d\tau} \psi_\tau(x) = \nu_\tau(\psi_\tau(x))$, where the learnable velocity field $\nu_\tau$ directs the transformation of each sample $X_0$ from a source distribution $p_0$, typically a Gaussian distribution, toward the target distribution $X_1 \sim p_1$ with $p_1$ representing the data distribution. If we prescribe the velocity field $\nu_\tau$ such that it guides every sample along a straight-line trajectory from $X_0$ to $X_1$, it is referred to as a rectified flow. In this case, $X_\tau$ represents the linear interpolation across the entire timespan between $X_0$ and $X_1$, which can be expressed as $X_\tau = \tau X_1+(1-\tau)X_0$. We employ a parameterized $\mathcal{M}_{\theta_g}$ to approximate $\nu_\tau$, leading to the following learning objective: $\mathcal{L}(\theta_g) = \mathbb{E}_{\tau, X_0, X_1} \left\| \mathcal{M}_{\theta_g} (X_\tau,\tau) - (X_1 - X_0) \right\|^2$. In our model, the rectified flow block takes the deterministic prediction $u_{t+1}$ from the composition of $L$ \ours blocks and the observations $\tilde{u}_{t}$ as conditioning inputs. Its objective is to generate the distribution of residuals $r_{t+1}=y-\sum_j u_{t+1}^j$, where $y$ is the ground truth for the solution at $t+1$. Our training loss then becomes:
\begin{align}
    \mathbb{E}_{X_0\sim\mathbf{N}(0,\mathbf{I}),\tau
    \sim (0,1)}[(\mathcal{M}_{\theta_g}( \tilde{u}_{t},\mathcal{F}_{\theta_d}(\tilde{u}_{t}),\tau,X_\tau) - (r_{t+1}-X_0))^2],
    \label{eq:fm}
\end{align}
where $X_\tau$ is the linear interpolation between the source sample $X_0$ and the target $r_{t+1}$.

  \begin{minipage}[t]{0.5\linewidth}
    \centering
    \scalebox{0.86}{
    \begin{algorithm}[H]
      \small
      \caption{\small{~StFT}} \label{alg:overview}
      \small{
      \textbf{Inputs: } history $\tilde u_t=(u_t,\dots,u_{t-k+1})$\\
      \textbf{Initialize: } blocks $l \in [1,L]$, patch sizes $p_{h_{l},w_{l}}$, truncation modes $m_{h_{l},w_{l}}$, overlaps $o_{h_{l},w_{l}}$, $u_{t+1}^0$ as $\mathrm{None}$, $u_{t+1}$ as $\mathbf0$\\
    $v\!\leftarrow\!\text{var idx}$\\
      \For{$l=1,\dots,L$}{
        $x_t \leftarrow \mathrm{TemporalStacking}(\tilde u_t,\,u_{t+1}^{l-1})$\;
        $\{x_{t}\}_v \leftarrow \mathrm{OLT}(x_t,\,p_{h,w},\,o_{h,w})$\;
        $\{x_{t+1}\}_v: $ \textbf{Invoke \emph{Freq.} \& \emph{S\text{-}t.} paths}\\
        $u_{t+1}^l \leftarrow \mathrm{OLDT}(\{x_{t+1}\}_v,\,p_{h,w},\,o_{h,w})$\\
        $u_{t+1} \leftarrow u_{t+1} + u_{t+1}^l$\\
      }
      \textbf{Return: } $u_{t+1}$
      }
    \end{algorithm}
    }
  \end{minipage}%
  \begin{minipage}[t]{0.5\linewidth}
    \centering
    \scalebox{0.81}{
    \begin{algorithm}[H]
      \small
      \caption{\small{~StFT block: \emph{Freq.} and \emph{S\text{-}t.} Paths}} \label{alg:stft_details}
      \small{
      \textbf{Frequency Path:} 
      $\bar x_{t,1}\!\leftarrow\!\mathrm{VariableMixer}^1(\{x_{t}\}_v)$
      $f_{t}\!\leftarrow\!\mathrm{FFTFilter}(\bar x_{t,1},m_h,m_w)$
      $f_{i,t}\!\leftarrow\!\mathrm{FreqEmbedder}(f_{i,t})$
      $f_{t}\!\leftarrow\!\mathrm{TransformerBlock}^1(f_{t})$
      $\{x_{t+1,1}\}_v\!\leftarrow\!\mathrm{iFFT}(f_{t},m_h,m_w)$
      $\{x_{t+1,1}\}_v\!\leftarrow\!\mathrm{Linear}^1(\{x_{t+1,1}\}_v)$
      
      \textbf{Spatiotemporal Path:} 
      $\bar x_{t,2}\!\leftarrow\!\mathrm{VariableMixer}^2(\{x_{t}\}_v)$
      $e_{t}\!\leftarrow\!\mathrm{StEmbedder}(\bar x_{t,2})$ 
      $e_{t}\!\leftarrow\!\mathrm{TransformerBlock}^2(e_{t})$ 
      $\{x_{t+1,2}\}_v\!\leftarrow\!\mathrm{Linear}^2(e_{t})$\\
      \textbf{Merge:} 
      $\{x_{t+1}\}_v \leftarrow \{x_{t+1,1}\}_v + \{x_{t+1,2}\}_v$
      }
    \end{algorithm}
    }
  \end{minipage}

\section{Experiments}
\subsection{Datasets}
In this section, we consider three spatio-temporal multi-physics systems arising from time-dependent PDEs of a variety of complexities, including a high-dimensional plasma dynamics system based on reconstructed equilibrium of DIII-D experimental discharges~\citep{bai2025transfer}, a 2D incompressible Navier-Stokes equation in velocity-pressure form within a square domain driven by an external force, and a viscous shallow-water equation modeling the dynamics of large-scale atmospheric flows on a spherical domain. The problem setup and data generation are detailed in Appendix~\ref{app1}.


\subsection{Experimental Setup}
\textbf{Long-term multi-physics prediction up to a horizon of 244 timesteps.} Our goal is to simulate long-time trajectories given a few initial observations. This task is particularly challenging due to the multiple correlated variables present in the Navier-Stokes and plasma magnetohydrodynamics (MHD), with the test trajectories consisting of snapshots that vary from $71$ to $244$. We employ an autoregressive framework for all the methods: during training, each model utilizes five historical snapshots to predict the next one in a forward pass. At test time, given the initial five snapshots of a trajectory, all models autoregressively generate the entire trajectories by iteratively predicting future states based on their own previous outputs.


\begin{figure}
\centering
\includegraphics[width=.47\columnwidth]
{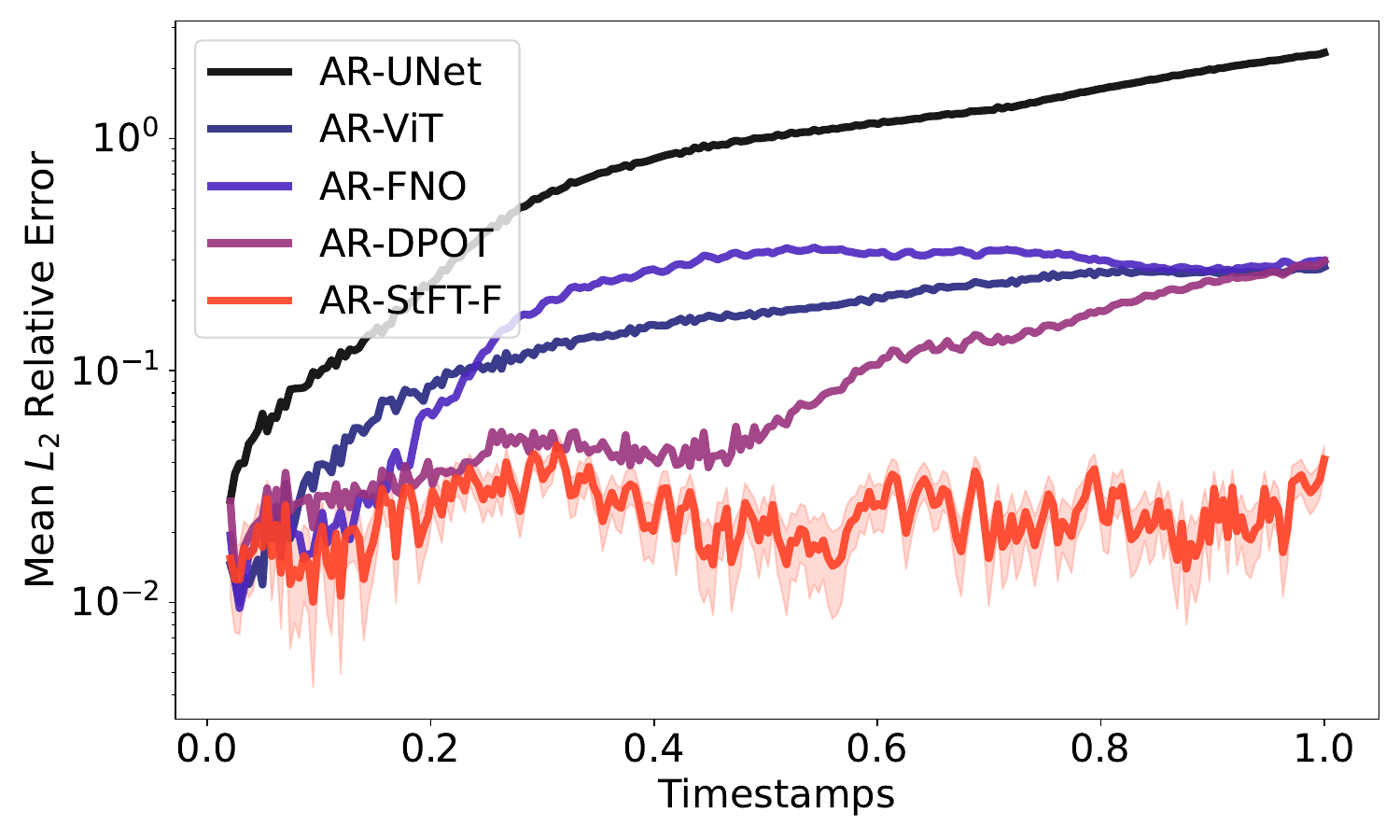}
\includegraphics[width=.495\columnwidth]
{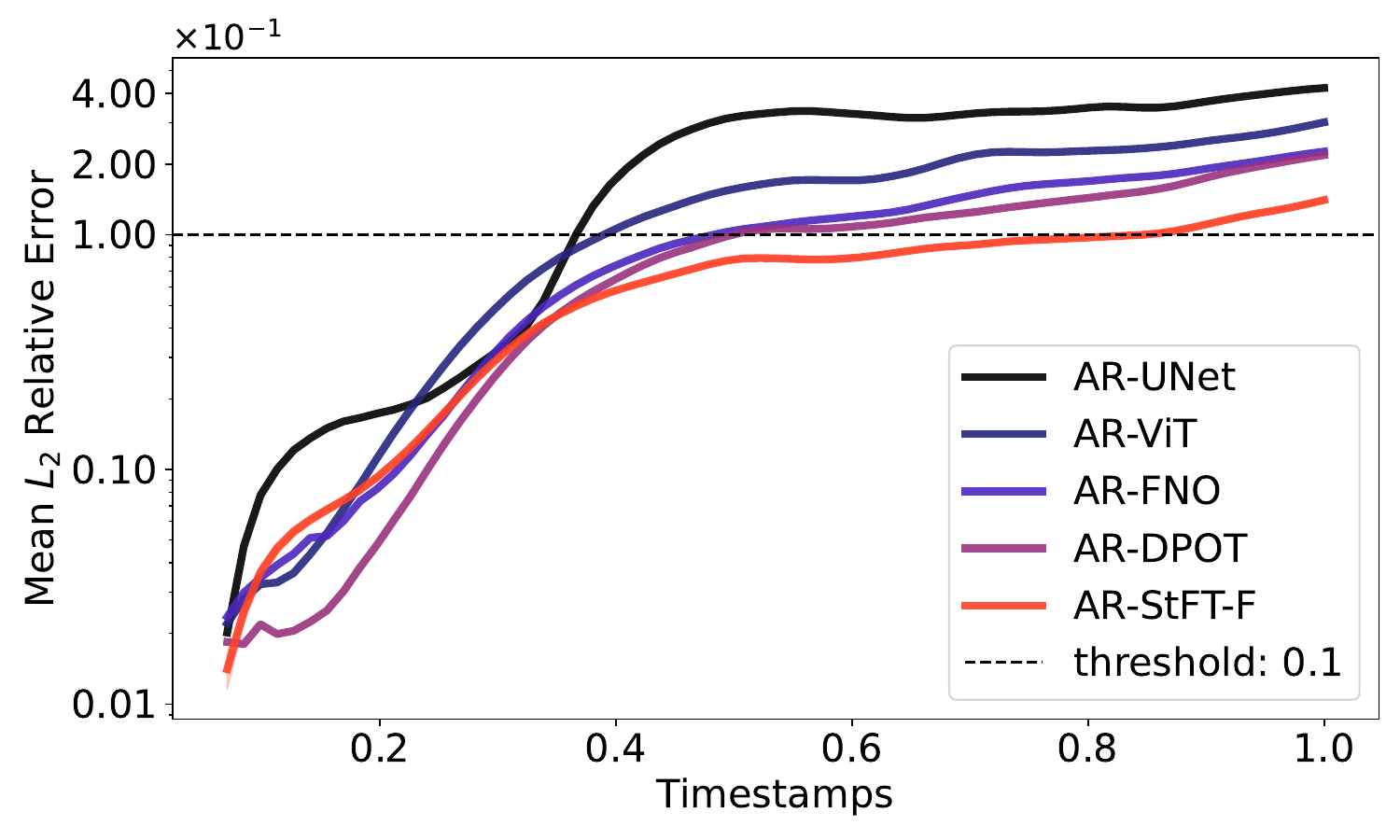}
\caption{\small Results of autoregressive prediction in $L_2$ relative error (log scale) across the timespan: (left)  perturbed parallel vector potential $\delta A_{\parallel}$ in plasma MHD, and (right) magnitude of velocity in shallow-water equation. The shaded region indicates the uncertainty distribution of $\sigma$ in the relative error of StFT-F. For a given error threshold, StFT-F maintains accuracy over at least twice the time horizon compared to the baselines.}
\label{fig:compare_error_versus_timestamp}
\end{figure}

\begin{table*}[t]
\centering
\scriptsize
\setlength{\tabcolsep}{3pt} 
\small
\caption{\small Quantitative results of our model and baselines in the same autoregressive framework: relative $L_2$ error over three spatiotemporal prediction systems. AR-\ours refers to the deterministic model's results. AR-StFT-F denotes the probabilistic model with residual refinement of StFT. All models have been subjected to hyperparameter tuning to ensure fair and optimal performance comparisons.}

\vspace{.12in}
\begin{tabular}{cccccccc}
\toprule
\textbf{Dataset} & \textbf{Variable(s)} & \textbf{AR-\ours}& \textbf{AR-StFT-F}& \textbf{AR-DPOT} & \textbf{AR-FNO}& \textbf{AR-ViT}& \textbf{AR-UNet$_b$}\\

\hline
\multirow{6}{*}{Plasma MHD}
& $\delta \phi$ &\Orange2.80e-2 & \NavyBlue{2.24e-2}& 1.04e-1& 2.28e-1&1.73e-1& 1.02e0 \\
& $\delta A_{\parallel}$ & \Orange2.45e-2 & \NavyBlue{2.30e-2}&8.36e-2 & 2.30e-1& 3.24e-1& 8.13e-1\\
& $\delta B_{\parallel}$ & \Orange3.05e-2& \NavyBlue{2.66e-2}&8.98e-2 & 2.33e-1& 1.95e-1& 7.79e-1\\
& $\delta n_e$ & \Orange2.84e-2& \NavyBlue{2.45e-2}& 8.64e-2& 2.33e-1& 2.08e-1& 1.01e0\\
& $\delta n_i$ &\Orange3.28e-2 & \NavyBlue{2.93e-2}&8.76e-2 & 2.33e-1& 2.18e-1& 1.04e0\\
& $\delta u_e$ & \Orange3.99e-2& \NavyBlue{3.73e-2}& 9.59e-2& 3.18e-1& 2.99e-1&6.96e-1\\
\hline
\multirow{3}{*}{Navier-Stokes} 
& $u$ &\Orange3.38e-2 &\NavyBlue{3.30e-2}&4.67e-2&  4.46e-2& 5.09e-2& 6.16e-2 \\
& $v$ & \Orange3.60e-2&\NavyBlue{3.17e-2}&4.52e-2&  4.57e-2& 4.60e-2& 6.15e-2 \\
& $p$ & \Orange5.16e-2&\NavyBlue{4.44e-2}&6.18e-2&  5.90e-2& 7.03e-2 & 7.84e-2\\
\hline
\multirow{1}{*}{Shallow-Water} 
& $\mathbf{V}$ &\NavyBlue{6.25e-2} & \Orange6.53e-2&7.97e-2 & 9.53e-2& 1.33e-1& 2.02e-1\\
\bottomrule
\end{tabular}
\label{tb:error}
\end{table*}

\textbf{Baseline setup.} We evaluate these datasets using the following well-known and state-of-the-art methods for comparison: autoregressive AFNO, autoregressive Fourier Neural Operator (AR-FNO), autoregressive U-Net (AR-UNet), and autoregressive vision transformer (AR-ViT). More specifically, for AFNO, we used the latest variant from the DPOT work ~\citep{hao2024dpot}, which is enhanced with a temporal aggregation layer and improved expressivity through the removal of enforced sparsity. For FNO, we use the authors' open-source implementation. For U-Net, we employ the implementation of the modified U-Net as evaluated in the recent BubbleML work~\citep{hassan2023bubbleml}, where the modified U-Net was initially used in PDEBench work ~\citep{takamoto2022pdebench}, and demonstrated superior performance over their baselines.

\textbf{Validation and hyperparameter tuning.} We divide the trajectories of each dataset into training, test, and validation sets. For each method, we identify the tunable hyperparameters, specify a range for each hyperparameter, and conduct a grid search. For our methods, we implement both StFT, the deterministic component of our model, and StFT-F, our model with a generative residual refinement block. We use the AdamW optimizer with a learning rate of $1e^{-4}$ to train those models on an A100 GPU. We ensure that all models are fairly and thoroughly trained by imposing an identical time budget across all models. More specifically, we impose 24 GPU hours on the plasma MHD and Navier-Stokes datasets, and 48 GPU hours on the shallow-water dataset. A comprehensive list of all the hyperparameters along with their respective ranges is provided in the Appendix~\ref{app3}.

\textbf{Evaluation metrics.} First, we evaluate the forecasting performance by calculating the mean $L_2$ relative error. For StFT-F, in order to obtain the mean prediction, we generate $50$ stochastic predictions at each autoregressive step, and then inject their mean into the next step. Additionally, we assess the uncertainty quantification capability of \ours-F by sampling $100$ trajectories for each test case. For each trajectory, at each autoregressive step, we generate a single prediction, which is subsequently fed into the next autoregressive step to iteratively forecast the full sequence. 

\subsection{Main Results}

\begin{figure}[htp]
\centering
\begin{minipage}[t]{0.42\textwidth}
    \vbox{
        \includegraphics[width=\linewidth,height=0.32\textheight,keepaspectratio]{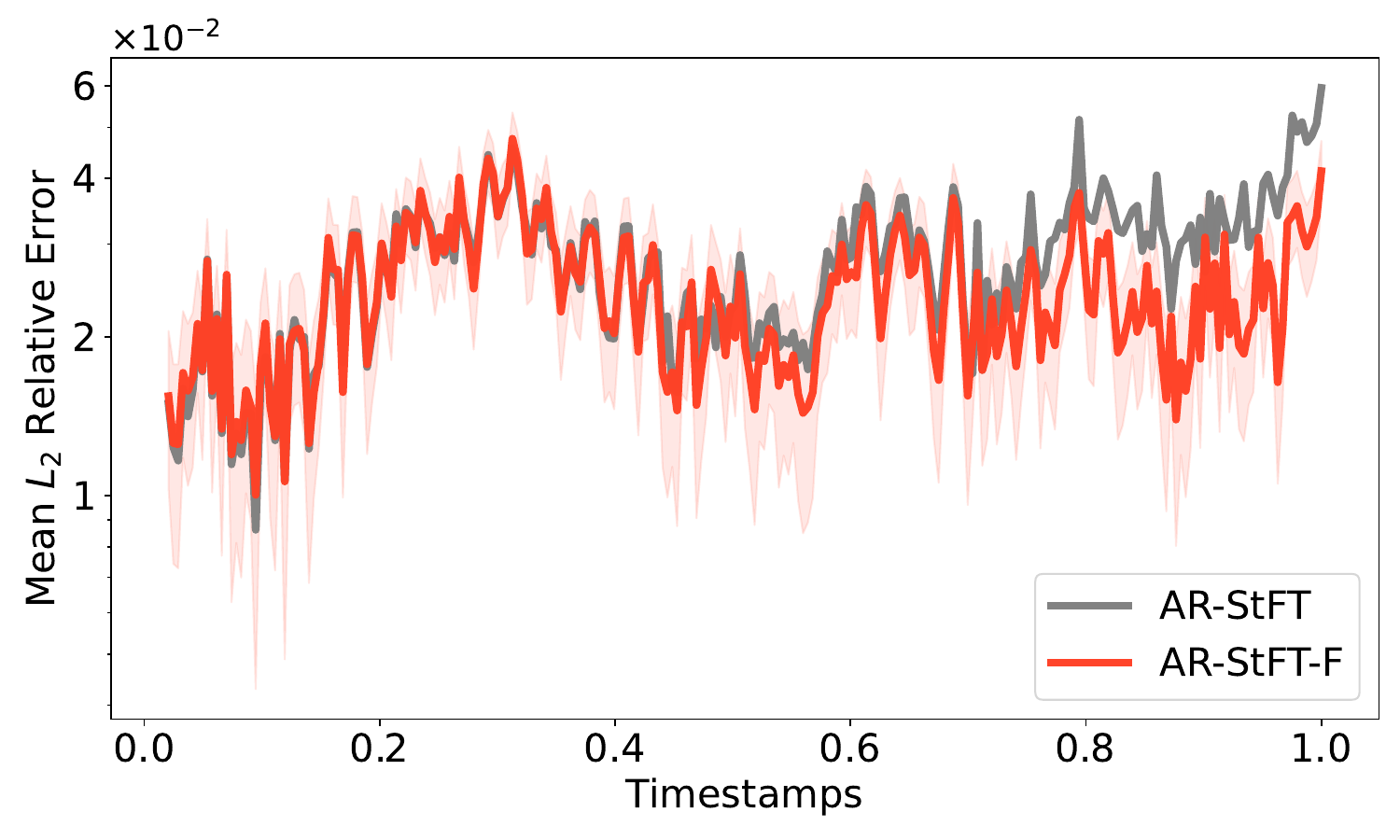}
        \vspace{0.02in}
        \includegraphics[width=\linewidth,height=0.32\textheight,keepaspectratio]{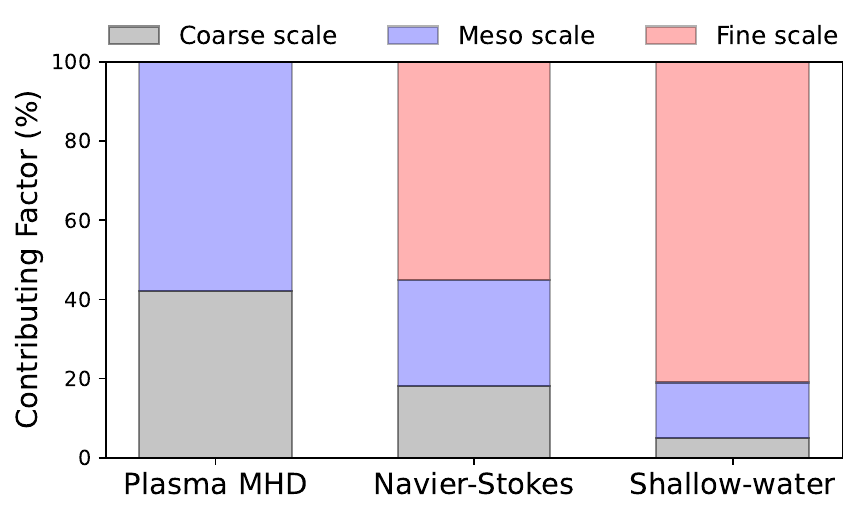}
    }
\end{minipage}%
\hfill
\begin{minipage}[t]{0.58\textwidth}
    \includegraphics[width=\linewidth,height=0.9\textheight,keepaspectratio]{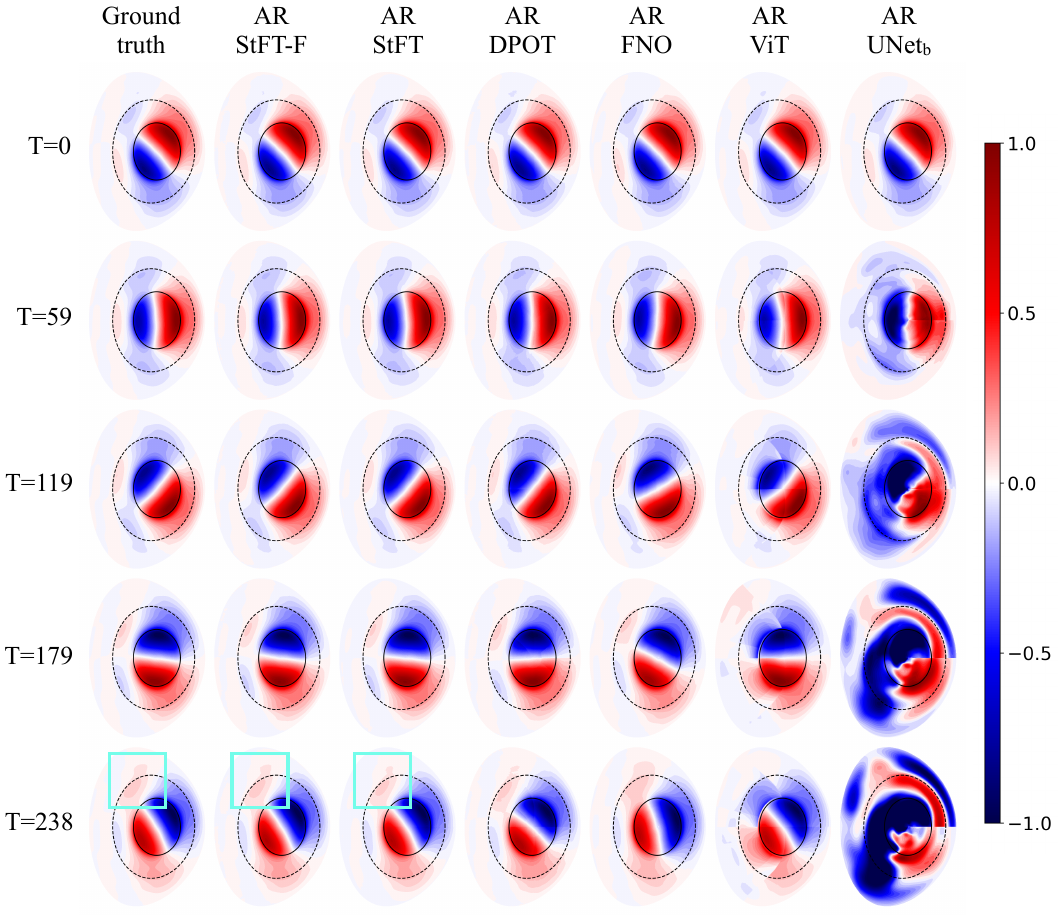}
\end{minipage}
\small
\caption{\small Left top: Comparing StFT and StFT-F in $L_2$ relative error across timestamps for two plasma MHD variables with uncertainty bands of StFT-F. Left bottom: The contribution of each \ours block. From bottom to top, the patch size decreases. For this plasma data, the first two levels are sufficient to capture the multi-scale structures, whereas for Navier-Stokes and shallow-water equations, the finest scale contributes more significantly. Right: Spatiotemporal evolution of perturbed electrostatic potential $\delta \phi$ predicted by autoregressive models. StFT and StFT-F remain accurate past $T=59$, whereas baseline methods exhibit out-of-phase predictions.}
\label{fig:compare_ours_error_versus_timestamp}
\end{figure}

Table~\ref{tb:error} shows the forecasting performance of all the models on the three applications. We present several test trajectories and visualize their uncertainties across the forecasting time horizon estimated by StFT-F, as illustrated in Figure~\ref{fig:uq}. 
\ours performs significantly better than all other baselines across all physical processes in the three applications. In the plasma MHD dataset, the test trajectory has a total of six coupled physics variables and $244$ snapshots. On average, \ours achieves a reduction in error by a factor of three compared to the best baseline, AR-DPOT(AFNO). Although AR-FNO maintains a high resolution in its long-term prediction, it fails to capture the correct dynamics of mode evolution, leading to out-of-phase predictions as shown in Figure~\ref{fig:compare_ours_error_versus_timestamp}.

We examine the error growth by plotting the $L_2$ relative errors over time, as illustrated in Figure~\ref{fig:compare_error_versus_timestamp} for several representative variables.
In the bottom figure for the shallow-water dataset, AR-FNO first appears to slightly better than all other methods during the short term from timestamp $0$ to $0.2$, and StFT-F shows superior performance soon after. For the plasma dataset, StFT-F demonstrates dominance starting from $t=0.2$ with a stable performance, while the errors of all baseline methods begin to propagate from that point onward, resulting in a rapid decline compared to StFT-F. Notably, StFT-F exhibits long-term stability relative to the other methods. 
As shown in Figure~\ref{fig:compare_ours_error_versus_timestamp}, both of our methods accurately capture the dominant mode phase, and StFT-F generates predictions that are more closely aligned with the ground truth compared to StFT.
We also compare the error over time comparing StFT and StFT-F. StFT-F begins to prevail from $t=0.6$. For the shallow-water and Navier-Stokes datasets, on average, \ours reduces the errors by an average of $27\%$ and $25\%$, respectively. In Navier-Stokes and plasma datasets, StFT-F not only surpasses \ours but also offers the additional capability of uncertainty quantification, achieving error reductions of $10\%$ in both cases. In the shallow-water dataset, we observe a slight increase in error with StFT-F. These results indicate that our method achieves superior long-term stability and accuracy among all other methods.
More visualizations and results are included in Appendix~\ref{app4}.

\subsection{Uncertainty Quantification}


\begin{wraptable}{r}{0.45\textwidth}
\centering
\vspace{-0pt}
\small
\caption{\small Average confidence interval (90\% and 95\%) coverage.}
\begin{tabular}{lcc}
\toprule
\textbf{Dataset} & \textbf{CI: 90\%} & \textbf{CI: 95\%} \\
\midrule
Plasma MHD    & 89.4\% & 92.5\% \\
Shallow-Water & 89.5\% & 98.0\%  \\
\bottomrule
\end{tabular}
\label{tab:ci}
\end{wraptable}

\begin{figure}
\vspace{-0.2in}
\centering
\includegraphics[width=\linewidth]{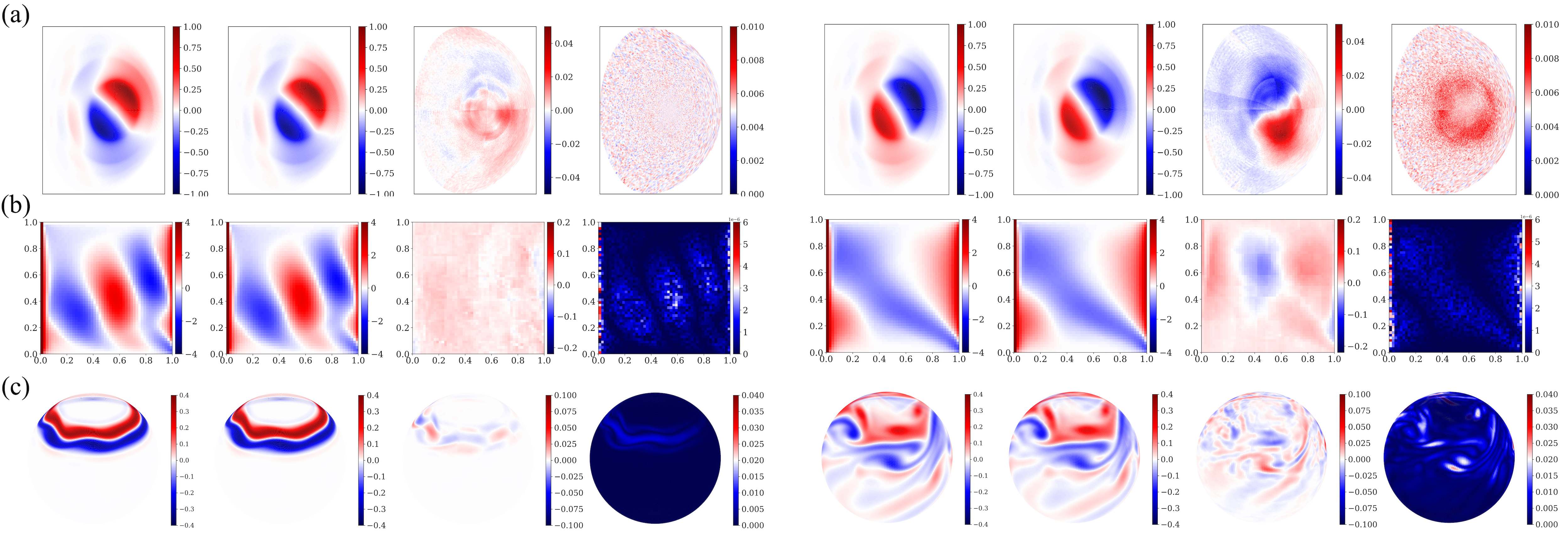}
\small
\caption{\small Evaluation of forecasting for three applications: ground truth, StFT-F prediction, residual, and uncertainty over time — shown for initial (left) and final (right) states. Variables include: (a) $\delta \phi$ in plasma MHD, (b) $u$ in Navier-Stokes, and (c) $\mathbf{V}$ in shallow-water equations.}
\label{fig:uq}
\end{figure}

Figure~\ref{fig:uq} presents the distribution of the empirical standard deviation along with the mean prediction. As observed, regions with large errors correspond to those exhibiting significant uncertainties predicted by StFT-F. Additionally, it is evident that uncertainties increase with time. This aligns with our expectation, as errors accumulate during the autoregressive forecasting process. Besides the empirical evaluation relying on the standard deviation plots, we further measure the robustness of StFT’s uncertainty quantification using confidence intervals as shown in Table ~\ref{tab:ci}. Specifically, we use the predicted uncertainty to compute empirical coverage by measuring the proportion of ground truth values that fall within a certain confidence interval around the predicted mean. This provides a more rigorous evaluation of StFT’s ability to capture uncertainties. Details about CI coverage and results regarding each variable are included in Appendix ~\ref{sec:UQ}. For the $90\%$ confidence interval, average coverage is very close to ideal value of $90\%$, demonstrating that StFT-F is well-calibrated around the $90\%$ confidence interval.

\subsection{Ablation Study}

\begin{wraptable}{r}{0.56\textwidth}
\centering
\footnotesize
\vspace{-10pt}
\setlength{\tabcolsep}{4pt}
\small
\caption{\small Effect of multi-scale structures and frequency path $\mathcal{F}$.}
\begin{tabular}{lcc}
\toprule
\textbf{Model} & \textbf{Plasma MHD} & \textbf{Shallow-Water} \\
\midrule
Mono-scale + $\mathcal{F}$ & 0.0805 / 0.105 & 2.5729 / 0.101 / 0.0975 \\
Multi-scale                 & 0.0404         & 0.0956 \\
Multi-scale + $\mathcal{F}$& \textbf{0.0307}  & \textbf{0.0625} \\
\bottomrule
\end{tabular}
\label{tb:ablation-a}
\vspace{-8pt}
\end{wraptable}





\textbf{Multi-scale and frequency path.} To evaluate the effectiveness of the hierarchical structure and the frequency path in StFT, we conduct an ablation study employing mono-scale versus multi-scale models, and employing models with and without the frequency path. Experimental results are shown in Table ~\ref{tb:ablation-a}.

\begin{wraptable}{r}{0.25\textwidth}
\centering
\footnotesize
\vspace{-15pt}
\setlength{\tabcolsep}{4pt}
\small
\caption{\small L2 relative error vs. number of scales (plasma MHD).}
\begin{tabular}{cc}
\toprule
\textbf{\# Scales} & \textbf{L2 Rel. Error} \\
\midrule
1 & 0.0805 \\
2 & \textbf{0.0307} \\
3 & 0.0385 \\
4 & 0.0391 \\
\bottomrule
\end{tabular}
\label{tb:ablation-b}
\end{wraptable}

\vspace{.1in}
\textbf{Convergence study on number of scales.} We evaluate the performance of StFT on the number of scales for the plasma MHD dataset. As shown in Table~\ref{tb:ablation-b}, it is observed that increasing the number of scales leads to a reduction in prediction error to a certain point, with the two-layer configuration achieving the lowest error. The inclusion of additional scales results in a slight error increase, indicating a convergence in model performance. Additionally, we conduct an ablation study on the overlapping tokenizer. More details are included in Appendix ~\ref{app2}.


\subsection{Additional Results}
\textbf{Computational and Space Complexity.}  We compare StFT with other baselines regarding the inference FLOPS per sample, training time and inference time per batch, and peak memory usage during training. We also compare StFT and StFT without the overlapping tokenizer, and we observed this enhancement comes with negligible impact on compute cost, as in Appendix ~\ref{computation}.

\textbf{Contribution of Each Scale.} To assess the contribution of each \ours block for a specific scale in fitting the training data, we quantify the weight of each block by: $\mathcal{W}_i = \frac{\| \mathbf{y}_i \|_2}{\| \mathbf{y} \|_2}$, where $\mathbf{y}_i$ represents the prediction from the $i$-th \ours block, and $\mathbf{y}$ denotes the ground truth. We normalize the contributions, and present the contributions in Figure~\ref{fig:compare_ours_error_versus_timestamp}. A greater contributing factor from the fine-scale layer in \ours is observed in the Navier-Stokes and shallow-water equations, attributed to the sharper changes and smaller scale structures inherent in the dynamics of higher-order nonlinearities.

{}


\section{Conclusion}
\vspace{-.1in}
In this paper, we propose a spatio-temporal Fourier transformer (StFT) for multi-scale and multi-physics long-term dynamics forecasting. Specifically, each StFT block is tailored to address a particular spatial scale, and through a hierarchical composition of multiple StFT blocks spanning different scales, StFT learns the interplay between multiple scales and interactions between multiple physical processes, resulting in stable and accurate long-term dynamics forecasting in an autoregressive manner. Furthermore, we propose and demonstrate the use of a generative residual correction mechanism, which enables meaningful quantification of uncertainties in the predictive model. 
Despite demonstrating superior forecasting ability in SciML, the model is based on regular grids, which constraints its applicability to irregular geometries. As part of future work, we plan to extend the framework to handle irregular domains, broadening its utility for more complex real-world scientific and engineering scenarios. The potential in improving its performance includes model parallelism across the multi-scale StFT blocks and extending StFT-F to end-to-end training.

\section*{Impact Statement}
Turbulence remains one of the great unsolved problems in physics.
Yet turbulence, whether driven by gravity, heating, or magnetic fields, manifests in physical phenomena spanning multi-scale fluid flow to plasma fusion reactors to planetary atmospheres to convective layers in stars to the flow of interstellar gas in stellar nurseries that span trillions of kilometers.
Where mathematics failed, data-driven machine learning models provide a pathway to understanding turbulence and gaining insights into not only the origins and ultimate fate of stars and planetary ecosystems, but also opens pathways to nearly infinite sources of clean energy.
Our StFT model presented in this paper is a stepping stone towards obviating the exponentially increasing computational costs of simulating geophysical systems (an inherently chaotic dynamical system requiring vast ensembles of high-resolution small time step simulations) as well as realization of digital twins to aid in the design and operational control of tokamak-based fusion power plants.

\section*{Acknowledgements}
ZB, DL, SW, and LO acknowledge support from the U.S. Department of Energy, Office of Science, SciDAC/Advanced Scientific Computing Research under Award Number DE-AC02-05CH11231 
with DL receiving additional support from the Margolis Foundation and MURI AFOSR under grant FA9550-20-1-035.
SZ was supported by MURI AFOSR grant FA9550-20-1-035, Margolis Foundation, and NSF Career Award IIS-2046295.
This research used resources of the National Energy Research Scientific Computing Center (NERSC), a U.S. Department of Energy Office of Science User Facility located at Lawrence Berkeley National Laboratory, operated under Contract Number DE-AC02-05CH11231,
as well as resources from the NCSA Delta GPU cluster from NCF Access program under award number IIS-2046295.

\bibliographystyle{plain}
\bibliography{neurips_2025}

\begin{thebibliography}{10}

\bibitem{albergo2022building}
Michael~S Albergo and Eric Vanden-Eijnden.
\newblock Building normalizing flows with stochastic interpolants.
\newblock {\em arXiv preprint arXiv:2209.15571}, 2022.

\bibitem{bai2025transfer}
Zhe Bai, Xishuo Wei, William Tang, Leonid Oliker, Zhihong Lin, and Samuel
  Williams.
\newblock Transfer learning nonlinear plasma dynamic transitions in low
  dimensional embeddings via deep neural networks.
\newblock {\em Machine Learning: Science and Technology}, 6(2):025015, 2025.

\bibitem{barlow2019north}
Mathew Barlow, William~J Gutowski, John~R Gyakum, Richard~W Katz, Young-Kwon
  Lim, Russ~S Schumacher, Michael~F Wehner, Laurie Agel, Michael Bosilovich,
  Allison Collow, et~al.
\newblock North american extreme precipitation events and related large-scale
  meteorological patterns: a review of statistical methods, dynamics, modeling,
  and trends.
\newblock {\em Climate Dynamics}, 53:6835--6875, 2019.

\bibitem{bi2023accurate}
Kaifeng Bi, Lingxi Xie, Hengheng Zhang, Xin Chen, Xiaotao Gu, and Qi~Tian.
\newblock Accurate medium-range global weather forecasting with 3d neural
  networks.
\newblock {\em Nature}, 619(7970):533--538, 2023.

\bibitem{brandstetter2022message}
Johannes Brandstetter, Daniel Worrall, and Max Welling.
\newblock Message passing neural pde solvers.
\newblock {\em arXiv preprint arXiv:2202.03376}, 2022.

\bibitem{burns2020dedalus}
Keaton~J Burns, Geoffrey~M Vasil, Jeffrey~S Oishi, Daniel Lecoanet, and
  Benjamin~P Brown.
\newblock Dedalus: A flexible framework for numerical simulations with spectral
  methods.
\newblock {\em Physical Review Research}, 2(2):023068, 2020.

\bibitem{cao2023lno}
Qianying Cao, Somdatta Goswami, and George~Em Karniadakis.
\newblock Lno: Laplace neural operator for solving differential equations.
\newblock {\em arXiv preprint arXiv:2303.10528}, 2023.

\bibitem{cao2021choose}
Shuhao Cao.
\newblock Choose a transformer: Fourier or galerkin.
\newblock {\em Advances in neural information processing systems},
  34:24924--24940, 2021.

\bibitem{chen1995universal}
Tianping Chen and Hong Chen.
\newblock Universal approximation to nonlinear operators by neural networks
  with arbitrary activation functions and its application to dynamical systems.
\newblock {\em IEEE transactions on neural networks}, 6(4):911--917, 1995.

\bibitem{cheung2011bayesian}
Sai~Hung Cheung, Todd~A Oliver, Ernesto~E Prudencio, Serge Prudhomme, and
  Robert~D Moser.
\newblock Bayesian uncertainty analysis with applications to turbulence
  modeling.
\newblock {\em Reliability Engineering \& System Safety}, 96(9):1137--1149,
  2011.

\bibitem{dosovitskiy2020image}
Alexey Dosovitskiy.
\newblock An image is worth 16x16 words: Transformers for image recognition at
  scale.
\newblock {\em arXiv preprint arXiv:2010.11929}, 2020.

\bibitem{esser2403scaling}
Patrick Esser, Sumith Kulal, Andreas Blattmann, Rahim Entezari, Jonas
  M{\"u}ller, Harry Saini, Yam Levi, Dominik Lorenz, Axel Sauer, Frederic
  Boesel, et~al.
\newblock Scaling rectified flow transformers for high-resolution image
  synthesis, 2024.
\newblock {\em URL https://arxiv. org/abs/2403.03206}, 1, 2024.

\bibitem{fan2021multiscale}
Haoqi Fan, Bo~Xiong, Karttikeya Mangalam, Yanghao Li, Zhicheng Yan, Jitendra
  Malik, and Christoph Feichtenhofer.
\newblock Multiscale vision transformers.
\newblock In {\em Proceedings of the IEEE/CVF international conference on
  computer vision}, pages 6824--6835, 2021.

\bibitem{graves2012control}
JP~Graves, IT~Chapman, S~Coda, M~Lennholm, M~Albergante, and M~Jucker.
\newblock Control of magnetohydrodynamic stability by phase space engineering
  of energetic ions in tokamak plasmas.
\newblock {\em Nature communications}, 3(1):624, 2012.

\bibitem{gupta2021multiwavelet}
Gaurav Gupta, Xiongye Xiao, and Paul Bogdan.
\newblock Multiwavelet-based operator learning for differential equations.
\newblock {\em Advances in neural information processing systems},
  34:24048--24062, 2021.

\bibitem{gupta2022towards}
Jayesh~K Gupta and Johannes Brandstetter.
\newblock Towards multi-spatiotemporal-scale generalized pde modeling.
\newblock {\em arXiv preprint arXiv:2209.15616}, 2022.

\bibitem{hao2024dpot}
Zhongkai Hao, Chang Su, Songming Liu, Julius Berner, Chengyang Ying, Hang Su,
  Anima Anandkumar, Jian Song, and Jun Zhu.
\newblock Dpot: Auto-regressive denoising operator transformer for large-scale
  pde pre-training.
\newblock {\em arXiv preprint arXiv:2403.03542}, 2024.

\bibitem{hao2023gnot}
Zhongkai Hao, Zhengyi Wang, Hang Su, Chengyang Ying, Yinpeng Dong, Songming
  Liu, Ze~Cheng, Jian Song, and Jun Zhu.
\newblock Gnot: A general neural operator transformer for operator learning.
\newblock In {\em International Conference on Machine Learning}, pages
  12556--12569. PMLR, 2023.

\bibitem{hassan2023bubbleml}
Sheikh Md~Shakeel Hassan, Arthur Feeney, Akash Dhruv, Jihoon Kim, Youngjoon
  Suh, Jaiyoung Ryu, Yoonjin Won, and Aparna Chandramowlishwaran.
\newblock Bubbleml: a multi-physics dataset and benchmarks for machine
  learning.
\newblock {\em arXiv preprint arXiv:2307.14623}, 2023.

\bibitem{ho2020denoising}
Jonathan Ho, Ajay Jain, and Pieter Abbeel.
\newblock Denoising diffusion probabilistic models.
\newblock {\em Advances in neural information processing systems},
  33:6840--6851, 2020.

\bibitem{ho2022video}
Jonathan Ho, Tim Salimans, Alexey Gritsenko, William Chan, Mohammad Norouzi,
  and David~J Fleet.
\newblock Video diffusion models.
\newblock {\em Advances in Neural Information Processing Systems},
  35:8633--8646, 2022.

\bibitem{jin2022mionet}
Pengzhan Jin, Shuai Meng, and Lu~Lu.
\newblock Mionet: Learning multiple-input operators via tensor product.
\newblock {\em SIAM Journal on Scientific Computing}, 44(6):A3490--A3514, 2022.

\bibitem{kong2020diffwave}
Zhifeng Kong, Wei Ping, Jiaji Huang, Kexin Zhao, and Bryan Catanzaro.
\newblock Diffwave: A versatile diffusion model for audio synthesis.
\newblock {\em arXiv preprint arXiv:2009.09761}, 2020.

\bibitem{kontolati2024learning}
Katiana Kontolati, Somdatta Goswami, George Em~Karniadakis, and Michael~D
  Shields.
\newblock Learning nonlinear operators in latent spaces for real-time
  predictions of complex dynamics in physical systems.
\newblock {\em Nature Communications}, 15(1):5101, 2024.

\bibitem{kramer2022probabilistic}
Nicholas Kr{\"a}mer, Jonathan Schmidt, and Philipp Hennig.
\newblock Probabilistic numerical method of lines for time-dependent partial
  differential equations.
\newblock In {\em International Conference on Artificial Intelligence and
  Statistics}, pages 625--639. PMLR, 2022.

\bibitem{kruger2024thinking}
SE~Kruger, Jarrod Leddy, EC~Howell, Sandeep Madireddy, C~Akcay,
  T~Bechtel~Amara, J~McClenaghan, LL~Lao, David Orozco, SP~Smith, et~al.
\newblock Thinking bayesian for plasma physicists.
\newblock {\em Physics of Plasmas}, 31(5), 2024.

\bibitem{li2023long}
Zhijie Li, Wenhui Peng, Zelong Yuan, and Jianchun Wang.
\newblock Long-term predictions of turbulence by implicit u-net enhanced
  fourier neural operator.
\newblock {\em Physics of Fluids}, 35(7), 2023.

\bibitem{li2022transformer}
Zijie Li, Kazem Meidani, and Amir~Barati Farimani.
\newblock Transformer for partial differential equations' operator learning.
\newblock {\em arXiv preprint arXiv:2205.13671}, 2022.

\bibitem{li2023fourier}
Zongyi Li, Daniel~Zhengyu Huang, Burigede Liu, and Anima Anandkumar.
\newblock Fourier neural operator with learned deformations for pdes on general
  geometries.
\newblock {\em Journal of Machine Learning Research}, 24(388):1--26, 2023.

\bibitem{li2020fourier}
Zongyi Li, Nikola Kovachki, Kamyar Azizzadenesheli, Burigede Liu, Kaushik
  Bhattacharya, Andrew Stuart, and Anima Anandkumar.
\newblock Fourier neural operator for parametric partial differential
  equations.
\newblock {\em arXiv preprint arXiv:2010.08895}, 2020.

\bibitem{lin2021operator}
Chensen Lin, Zhen Li, Lu~Lu, Shengze Cai, Martin Maxey, and George~Em
  Karniadakis.
\newblock Operator learning for predicting multiscale bubble growth dynamics.
\newblock {\em The Journal of Chemical Physics}, 154(10), 2021.

\bibitem{lipman2022flow}
Yaron Lipman, Ricky~TQ Chen, Heli Ben-Hamu, Maximilian Nickel, and Matt Le.
\newblock Flow matching for generative modeling.
\newblock {\em arXiv preprint arXiv:2210.02747}, 2022.

\bibitem{lippe2023pde}
Phillip Lippe, Bas Veeling, Paris Perdikaris, Richard Turner, and Johannes
  Brandstetter.
\newblock Pde-refiner: Achieving accurate long rollouts with neural pde
  solvers.
\newblock {\em Advances in Neural Information Processing Systems},
  36:67398--67433, 2023.

\bibitem{liu2022rectified}
Qiang Liu.
\newblock Rectified flow: A marginal preserving approach to optimal transport.
\newblock {\em arXiv preprint arXiv:2209.14577}, 2022.

\bibitem{liu2022flow}
Xingchao Liu, Chengyue Gong, and Qiang Liu.
\newblock Flow straight and fast: Learning to generate and transfer data with
  rectified flow.
\newblock {\em arXiv preprint arXiv:2209.03003}, 2022.

\bibitem{liu2022ht}
Xinliang Liu, Bo~Xu, and Lei Zhang.
\newblock Ht-net: Hierarchical transformer based operator learning model for
  multiscale pdes.
\newblock 2022.

\bibitem{liu2021swin}
Ze~Liu, Yutong Lin, Yue Cao, Han Hu, Yixuan Wei, Zheng Zhang, Stephen Lin, and
  Baining Guo.
\newblock Swin transformer: Hierarchical vision transformer using shifted
  windows.
\newblock In {\em Proceedings of the IEEE/CVF international conference on
  computer vision}, pages 10012--10022, 2021.

\bibitem{long2024arbitrarily}
Da~Long, Zhitong Xu, Guang Yang, Akil Narayan, and Shandian Zhe.
\newblock Arbitrarily-conditioned multi-functional diffusion for multi-physics
  emulation.
\newblock {\em arXiv preprint arXiv:2410.13794}, 2024.

\bibitem{lu2021learning}
Lu~Lu, Pengzhan Jin, Guofei Pang, Zhongqiang Zhang, and George~Em Karniadakis.
\newblock Learning nonlinear operators via deeponet based on the universal
  approximation theorem of operators.
\newblock {\em Nature machine intelligence}, 3(3):218--229, 2021.

\bibitem{lucor2003predictability}
D~Lucor, D~Xiu, C-H Su, and GE~Karniadakis.
\newblock Predictability and uncertainty in cfd.
\newblock {\em International Journal for Numerical Methods in Fluids},
  43(5):483--505, 2003.

\bibitem{mccabe2023towards}
Michael McCabe, Peter Harrington, Shashank Subramanian, and Jed Brown.
\newblock Towards stability of autoregressive neural operators.
\newblock {\em arXiv preprint arXiv:2306.10619}, 2023.

\bibitem{natrajan2007statistical}
Vinay~K Natrajan, Eiichiro Yamaguchi, and Kenneth~T Christensen.
\newblock Statistical and structural similarities between micro-and macroscale
  wall turbulence.
\newblock {\em Microfluidics and Nanofluidics}, 3:89--100, 2007.

\bibitem{pathak2022fourcastnet}
Jaideep Pathak, Shashank Subramanian, Peter Harrington, Sanjeev Raja, Ashesh
  Chattopadhyay, Morteza Mardani, Thorsten Kurth, David Hall, Zongyi Li, Kamyar
  Azizzadenesheli, et~al.
\newblock Fourcastnet: A global data-driven high-resolution weather model using
  adaptive fourier neural operators.
\newblock {\em arXiv preprint arXiv:2202.11214}, 2022.

\bibitem{peebles2023scalable}
William Peebles and Saining Xie.
\newblock Scalable diffusion models with transformers.
\newblock In {\em Proceedings of the IEEE/CVF International Conference on
  Computer Vision}, pages 4195--4205, 2023.

\bibitem{peters2009multiscale}
Norbert Peters.
\newblock Multiscale combustion and turbulence.
\newblock {\em Proceedings of the Combustion Institute}, 32(1):1--25, 2009.

\bibitem{polyak2024movie}
Adam Polyak, Amit Zohar, Andrew Brown, Andros Tjandra, Animesh Sinha, Ann Lee,
  Apoorv Vyas, Bowen Shi, Chih-Yao Ma, Ching-Yao Chuang, et~al.
\newblock Movie gen: A cast of media foundation models.
\newblock {\em arXiv preprint arXiv:2410.13720}, 2024.

\bibitem{prasthofer2022variable}
Michael Prasthofer, Tim De~Ryck, and Siddhartha Mishra.
\newblock Variable-input deep operator networks.
\newblock {\em arXiv preprint arXiv:2205.11404}, 2022.

\bibitem{rahman2022u}
Md~Ashiqur Rahman, Zachary~E Ross, and Kamyar Azizzadenesheli.
\newblock U-no: U-shaped neural operators.
\newblock {\em arXiv preprint arXiv:2204.11127}, 2022.

\bibitem{ruhling2023dyffusion}
Salva R{\"u}hling~Cachay, Bo~Zhao, Hailey Joren, and Rose Yu.
\newblock Dyffusion: A dynamics-informed diffusion model for spatiotemporal
  forecasting.
\newblock {\em Advances in neural information processing systems},
  36:45259--45287, 2023.

\bibitem{scher2018predicting}
Sebastian Scher and Gabriele Messori.
\newblock Predicting weather forecast uncertainty with machine learning.
\newblock {\em Quarterly Journal of the Royal Meteorological Society},
  144(717):2830--2841, 2018.

\bibitem{seo2024avoiding}
Jaemin Seo, SangKyeun Kim, Azarakhsh Jalalvand, Rory Conlin, Andrew Rothstein,
  Joseph Abbate, Keith Erickson, Josiah Wai, Ricardo Shousha, and Egemen
  Kolemen.
\newblock Avoiding fusion plasma tearing instability with deep reinforcement
  learning.
\newblock {\em Nature}, 626(8000):746--751, 2024.

\bibitem{singer2022make}
Uriel Singer, Adam Polyak, Thomas Hayes, Xi~Yin, Jie An, Songyang Zhang, Qiyuan
  Hu, Harry Yang, Oron Ashual, Oran Gafni, et~al.
\newblock Make-a-video: Text-to-video generation without text-video data.
\newblock {\em arXiv preprint arXiv:2209.14792}, 2022.

\bibitem{song2020score}
Yang Song, Jascha Sohl-Dickstein, Diederik~P Kingma, Abhishek Kumar, Stefano
  Ermon, and Ben Poole.
\newblock Score-based generative modeling through stochastic differential
  equations.
\newblock {\em arXiv preprint arXiv:2011.13456}, 2020.

\bibitem{tadmor2012review}
Eitan Tadmor.
\newblock A review of numerical methods for nonlinear partial differential
  equations.
\newblock {\em Bulletin of the American Mathematical Society}, 49(4):507--554,
  2012.

\bibitem{takamoto2022pdebench}
Makoto Takamoto, Timothy Praditia, Raphael Leiteritz, Daniel MacKinlay,
  Francesco Alesiani, Dirk Pfl{\"u}ger, and Mathias Niepert.
\newblock Pdebench: An extensive benchmark for scientific machine learning.
\newblock {\em Advances in Neural Information Processing Systems},
  35:1596--1611, 2022.

\bibitem{tian2024transfusion}
Sibo Tian, Minghui Zheng, and Xiao Liang.
\newblock Transfusion: A practical and effective transformer-based diffusion
  model for 3d human motion prediction.
\newblock {\em IEEE Robotics and Automation Letters}, 2024.

\bibitem{tran2021factorized}
Alasdair Tran, Alexander Mathews, Lexing Xie, and Cheng~Soon Ong.
\newblock Factorized fourier neural operators.
\newblock {\em arXiv preprint arXiv:2111.13802}, 2021.

\bibitem{vaswani2017attention}
A~Vaswani.
\newblock Attention is all you need.
\newblock {\em Advances in Neural Information Processing Systems}, 2017.

\bibitem{voleti2022mcvd}
Vikram Voleti, Alexia Jolicoeur-Martineau, and Chris Pal.
\newblock Mcvd-masked conditional video diffusion for prediction, generation,
  and interpolation.
\newblock {\em Advances in neural information processing systems},
  35:23371--23385, 2022.

\bibitem{von1974studies}
Stodiek von Goeler, W~Stodiek, and N~Sauthoff.
\newblock Studies of internal disruptions and m= 1 oscillations in tokamak
  discharges with soft—x-ray tecniques.
\newblock {\em Physical Review Letters}, 33(20):1201, 1974.

\bibitem{wang2006large}
Bin Wang, Brian Hoskins, and Bin Wang.
\newblock Large-scale atmospheric dynamics.
\newblock {\em The Asian Monsoon}, pages 357--415, 2006.

\bibitem{NEURIPS2024_7f605d59}
Qi~Wang, Pu~Ren, Hao Zhou, Xin-Yang Liu, Zhiwen Deng, Yi~Zhang, Ruizhi Chengze,
  Hongsheng Liu, Zidong Wang, Jian-Xun Wang, Ji-Rong Wen, Hao Sun, and Yang
  Liu.
\newblock P\^{}2c\^{}2net: Pde-preserved coarse correction network for
  efficient prediction of spatiotemporal dynamics.
\newblock In A.~Globerson, L.~Mackey, D.~Belgrave, A.~Fan, U.~Paquet,
  J.~Tomczak, and C.~Zhang, editors, {\em Advances in Neural Information
  Processing Systems}, volume~37, pages 68897--68925. Curran Associates, Inc.,
  2024.

\bibitem{wang2024bridging}
Sifan Wang, Jacob~H Seidman, Shyam Sankaran, Hanwen Wang, George~J Pappas, and
  Paris Perdikaris.
\newblock Bridging operator learning and conditioned neural fields: A unifying
  perspective.
\newblock {\em arXiv preprint arXiv:2405.13998}, 2024.

\bibitem{wang2021learning}
Sifan Wang, Hanwen Wang, and Paris Perdikaris.
\newblock Learning the solution operator of parametric partial differential
  equations with physics-informed deeponets.
\newblock {\em Science advances}, 7(40):eabi8605, 2021.

\bibitem{wang2022improved}
Sifan Wang, Hanwen Wang, and Paris Perdikaris.
\newblock Improved architectures and training algorithms for deep operator
  networks.
\newblock {\em Journal of Scientific Computing}, 92(2):35, 2022.

\bibitem{wolleb2022diffusion}
Julia Wolleb, Florentin Bieder, Robin Sandk{\"u}hler, and Philippe~C Cattin.
\newblock Diffusion models for medical anomaly detection.
\newblock In {\em International Conference on Medical image computing and
  computer-assisted intervention}, pages 35--45. Springer, 2022.

\bibitem{zhang2022nested}
Zizhao Zhang, Han Zhang, Long Zhao, Ting Chen, Sercan~{\"O} Arik, and Tomas
  Pfister.
\newblock Nested hierarchical transformer: Towards accurate, data-efficient and
  interpretable visual understanding.
\newblock In {\em Proceedings of the AAAI Conference on Artificial
  Intelligence}, volume~36, pages 3417--3425, 2022.

\end{thebibliography}

\newpage
\appendix
\onecolumn

\section{Problem Setup and Datasets}\label{app1}

\subsection{Plasma magnetohydrodynamic (MHD) equations} 
We consider magnetohydrodynamic (MHD) equations that characterize the plasma instabilities in fusion tokamaks. The coupled multi-physics system includes the continuity equation solving charge density $\delta n$, Poisson's equation solving $\delta\phi$, the Ampere's law to solving $\delta u_\parallel$, the Faraday's law with the assumption $E_\parallel=0$ to solving $\delta A_\parallel$, and the perpendicular force balance equation to solving $\delta B_\parallel$.
The first continuity equation for gyrocenter charge density is expressed as,
\begin{align}
\begin{split}
 & \frac{\partial\delta n}{\partial t}+\mathbf{B}_{0}\cdot\nabla\left(\frac{n_{0}\delta u_{\parallel}}{B_{0}}\right)-n_{0}\mathbf{v}_{*}\cdot\frac{\nabla B_{0}}{B_{0}}+\delta\mathbf{B}_{\perp}\cdot\nabla\left(\frac{n_{0}u_{\parallel0}}{B_{0}}\right)\\
 & -\frac{\nabla\times\mathbf{B_{0}}}{eB_{0}^{2}}\cdot\left(\nabla\delta P_{\parallel}+\frac{\left(\delta P_{\perp}-\delta P_{\parallel}\right)\nabla B_{0}}{B_{0}}\right)\\
 & +\nabla\cdot\left(\frac{\delta P_{\parallel}\mathbf{b}_{0}\nabla\times\mathbf{b}_{0}\cdot\mathbf{b}_{0}}{eB_{0}}\right) -\frac{\mathbf{b}_{0}\times\nabla\delta B_{\parallel}}{e}\cdot\nabla\left(\frac{P_{0}}{B_{0}^{2}}\right)\\
 &-\frac{\nabla\times\mathbf{b}_{0}\cdot\nabla\delta B_{\parallel}}{eB_{0}^{2}}P_{0}=0,\label{eq:MHD_continuity}
\end{split}
\end{align}
where $n$ is the density, $B$ is the magnetic field, $u_\parallel$ is the parallel flow velocity, and $P$ is the pressure. The perturbed quantities are denoted by $\delta$ with the equilibrium states including temperature, density, magnetic field and the flux surface from the reconstruction of DIII-D experiments. Here, $\delta n=\delta n_e + q_i\delta n_i/q_e$ stands for the difference of ion and electron density, and $\delta u_\parallel=\delta u_{\parallel e} +q_i \delta u_{\parallel i}/q_e$ denotes the difference of ion and electron flow. We have $\mathbf{v}_{*} = \mathbf{b}_{0}\times\nabla\left(\delta P_{\parallel}+\delta P_{\perp}\right)/\left(n_{0}m_{e}\Omega_{e}\right)$, where $m_e$ is the electron mass, and $\Omega_e=eB_0/m_e$ is the electron cyclotron frequency. The perturbed electron parallel flow $\delta u_\parallel$ can be solved from Ampere's law,

\begin{equation}
    \delta u_{\parallel}=\frac{1}{\mu_{0}en_{0}}\nabla_{\perp}^{2}\delta A_{\parallel},\label{eq:Ampere_MHD}
\end{equation}
where $\mu_0$ is the permeability of vacuum. $\delta A_\parallel$ is the perturbed vector potential. In the single fluid model, $E_\parallel = 0$ is assumed. Then $\delta A_\parallel$ can be solved from
\begin{equation}
    \frac{\partial A_{\parallel}}{\partial t}=\mathbf{b}_{0}\cdot\nabla\phi,\label{eq:Apara_MHD}
\end{equation}
and the electrostatic potential $\phi$ can be solved from gyrokinetic Poisson's equation (the quasi-neutrality condition)
\begin{equation}
    \frac{c^{2}}{v_{A}^{2}}\nabla_{\perp}^{2}\phi=\frac{e\delta n}{\epsilon_{0}},\label{eq:poisson_MHD}
\end{equation}
 where $c$ is the speed of light, $v_{A}$ is the Alfv\'{e}n velocity, and $\epsilon_{0}$ is the dielectric constant of vacuum. The parallel magnetic perturbation $\delta B$ is given by the perpendicular force balance,
\begin{equation}
\frac{\delta B_{\parallel}}{B_{0}}=-\frac{\beta_e}{2}\frac{\delta P_\perp}{P_{0}}=-\frac{\beta_{e}}{2}\frac{\partial P_{0}}{\partial\psi_{0}}\frac{\delta\psi}{P_{0}}.\label{eq:bpara_MHD}
\end{equation}
The perturbed pressure in the fluid limit can be calculated by
\begin{equation}
\begin{aligned}
\delta P_{\perp} & =\frac{\partial P_{0}}{\partial\psi_{0}}\delta\psi-2\frac{\delta B_{\parallel}}{B_{0}}P_{0},\\
\delta P_{\parallel} & =\frac{\partial P_{0}}{\partial\psi_{0}}\delta\psi-\frac{\delta B_{\parallel}}{B_{0}}P_{0}.\label{eq:Ppara_ad_MHD}
\end{aligned}
\end{equation}
In these equations, $\psi_0$ and $\delta\psi$ is the equilibrium and perturbed magnetic flux, and the evolution of $\delta\psi$ is solved from
\begin{align}
    \frac{\partial\delta\psi}{\partial t} & =-\frac{\partial\phi}{\partial\alpha_{0}},\label{eq:dpsi}
\end{align}
where $\alpha_0$ is from the Clebsch representation of $\mathbf{B}$ field, and $\mathbf{B}_0=\nabla\psi_0\times\nabla\alpha_0$. We run a linear gyrokinetic simulation with a $100\times250\times24$ mesh in radial, poloidal and parallel directions. The time step is set to $\Delta t=0.005R_0/C_s = 1.483\times 10^{-8} \text{s}$. We keep both $n=0, 1$ modes, generate a trajectory of $128,000$ time steps, and collect the data every $100$ snapshots. We focus on emulating the dynamics of electrostatic potential $\delta\phi$, parallel vector potential $\delta A_{\parallel}$, electron number density $\delta n_e$, ion number density $\delta n_i$, and electron velocity $\delta u_e$ in their trajectories.

\begin{figure*}
\centering
\includegraphics[width=.495\linewidth]{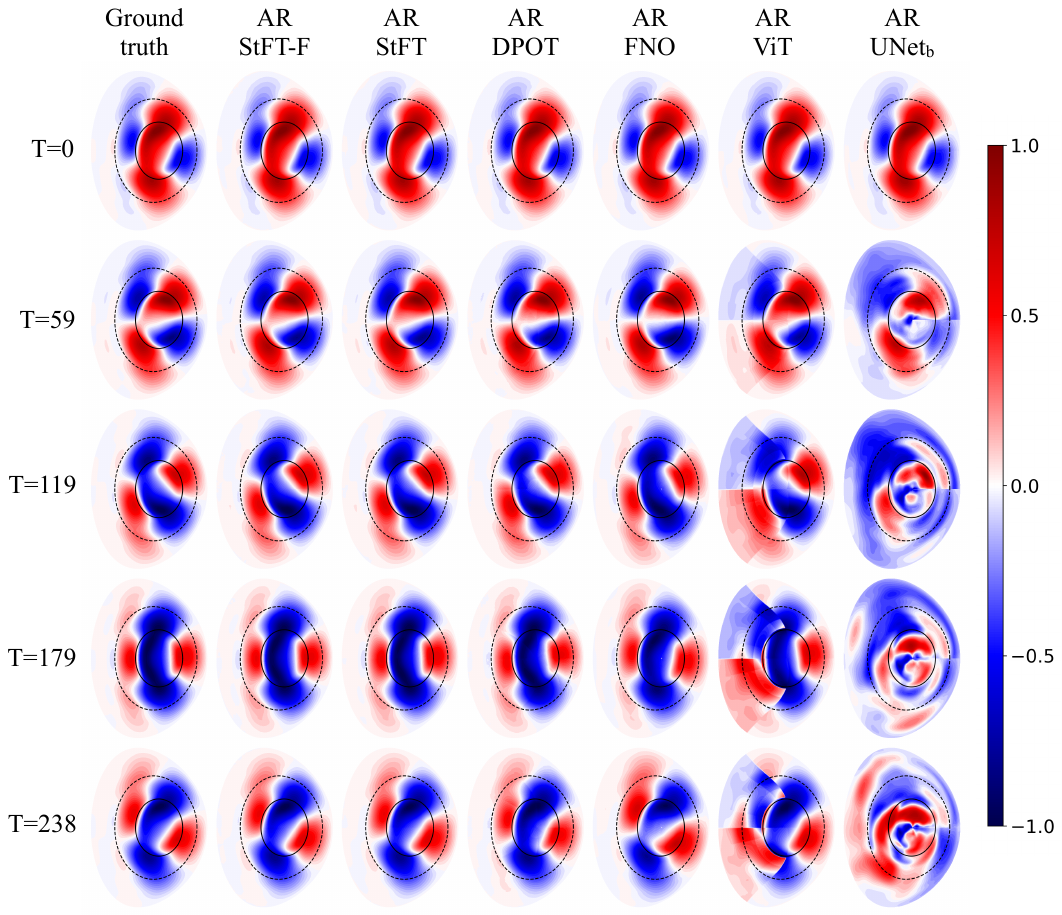}
\includegraphics[width=.495\linewidth]
{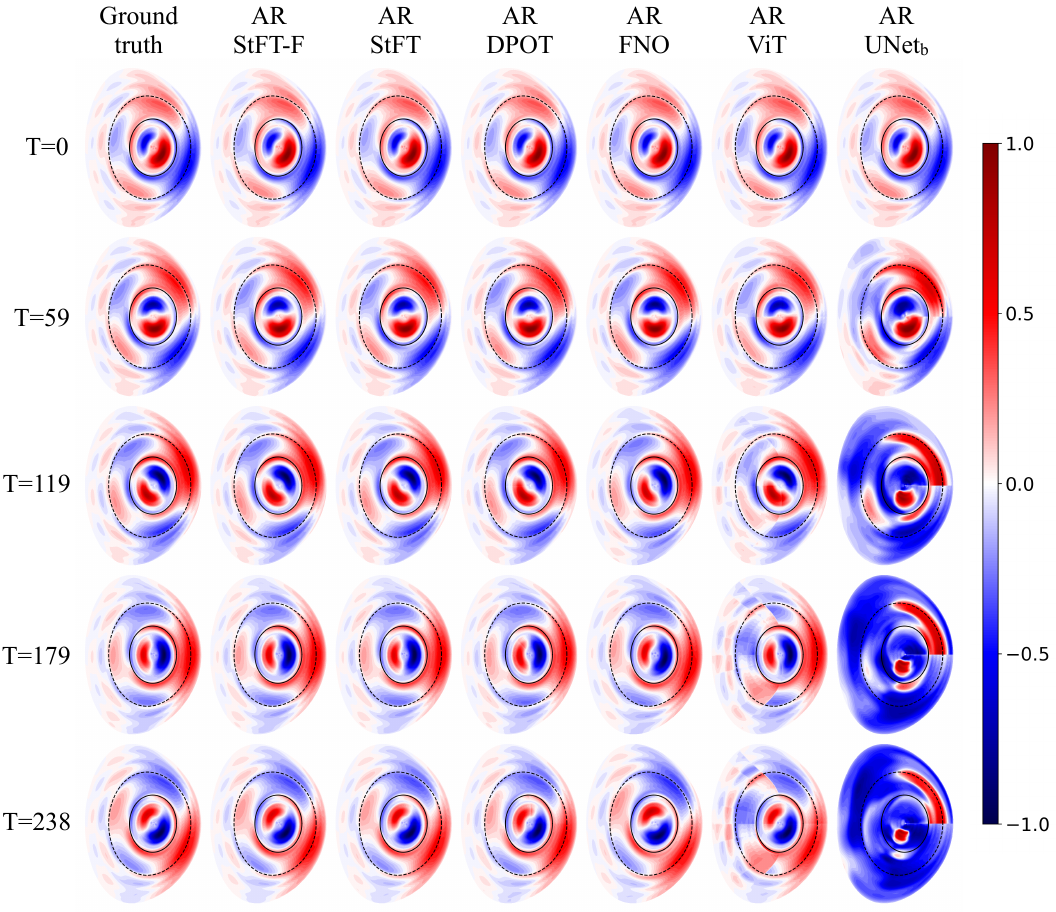}
\caption{\small Temporal evolution of normalized perturbed parallel vector potential $\delta A_{\parallel}$ and perturbed electron density $\delta n_e$ contours predicted by different models: StFT-F, StFT, FNO, ViT and U-Net. Significant phase differences between the predictions of the models appear after $T=59$, where StFT and StFT-F perform stable across the forecasting time horizon.} 
\label{fig:vis_plasma}
\end{figure*}

\subsection{2D incompressible Navier-Stokes equations}

We consider the 2D incompressible Navier-Stokes (NS) equation on a rectangular domain $(x, y) \in [0, 1]^2$,
\begin{equation}
\begin{aligned}
    \frac{\partial u}{\partial t} + \frac{\partial p}{\partial x} &= -u\frac{\partial u}{\partial x} - v\frac{\partial u}{\partial y} + \frac{1}{\text{Re}} \nabla^2 u +f(x,y), \\
    \frac{\partial v}{\partial t} + \frac{\partial p}{\partial y} &= -u\frac{\partial v}{\partial x} - v\frac{\partial v}{\partial y} + \frac{1}{\text{Re}} \nabla^2 v +f(x,y),\\
    \frac{\partial u}{\partial x} + \frac{\partial v}{\partial y} &= 0,
\end{aligned}
\end{equation}
where $u$ and $v$ represent the velocity components in the x and y directions, and $p$ represents the pressure. $f(x,y)$ is the source term, and we set it to $e^{-100((x-0.5)^2 + (y-0.5)^2)}$. The Reynolds number is set to 1000. We run a finite difference solver to compute the solutions on a $50\times50$ spatial grid, with the temporal domain discretized into a total of 101 timestamps over $T\in[0,20]$. We generated a total of 100 trajectories by sampling the four boundary conditions uniformly from $(0.1,0.6)$.

\begin{figure*}
\centering
\includegraphics[width=.49\linewidth]{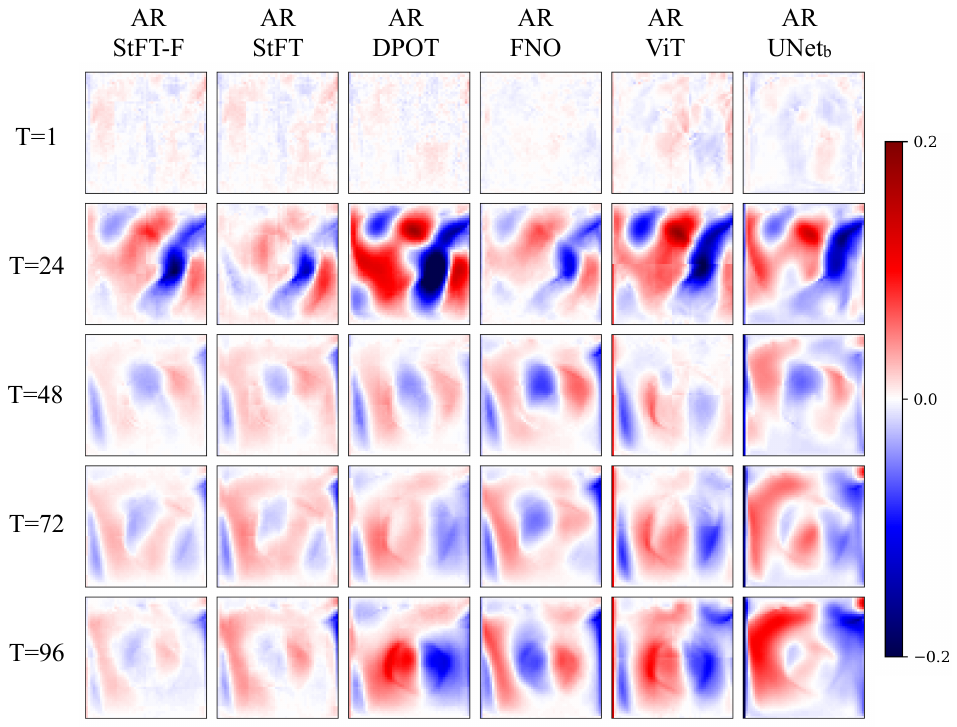}
\includegraphics[width=.5\linewidth]
{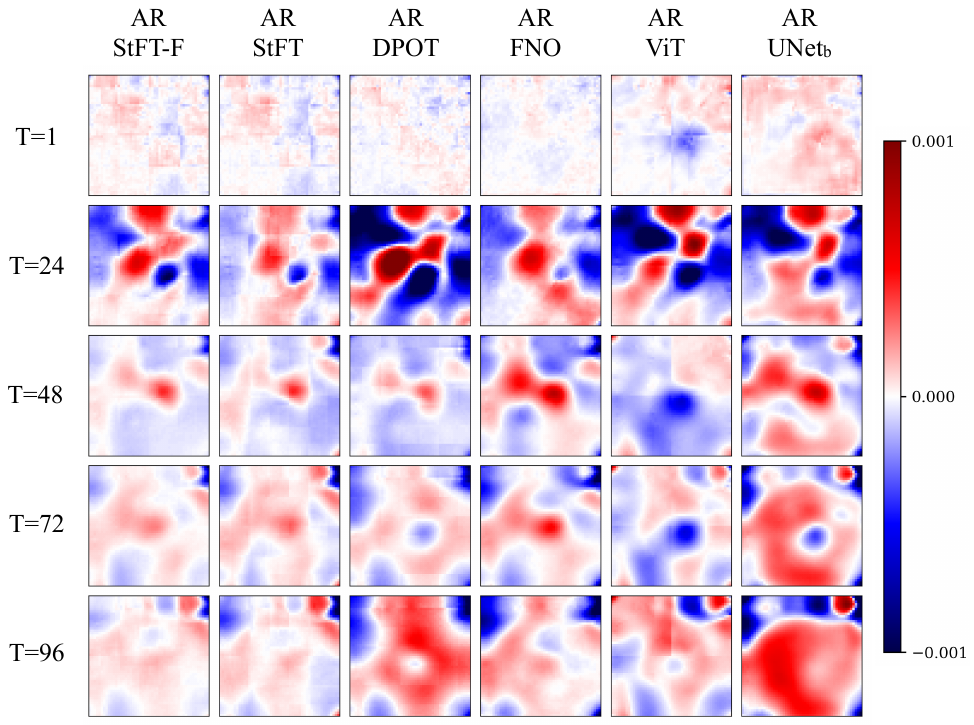}
\caption{\small 2D incompressible Navier-Stokes equation: pointwise error of the predicted evolution of velocity component $u$ and pressure $p$ contours across different models: StFT-F, StFT, FNO, ViT and U-Net. For long-term predictions, StFT and StFT-F demonstrate lower residuals compared to other models.} 
\label{fig:vis_ns}
\end{figure*}

\subsection{Spherical shallow-water equations}
We consider the viscous shallow-water equations modeling the dynamics of large-scale atmospheric flows:
\begin{equation}
\begin{aligned}
\frac{D\mathbf{V}}{Dt} &= -f \mathbf{k} \times \mathbf{V} - g \nabla h + \nu \nabla^2 \mathbf{V}, \\
\frac{Dh}{Dt} &= -h \nabla \cdot \mathbf{V} + \nu \nabla^2 h, \quad x \in \Omega, \; t \in [0, 1],
\end{aligned}
\end{equation}
where $\mathbf{V}$ is the velocity vector tangential to the spherical surface, $\mathbf{k}$ is the unit vector normal to the surface, $h$ is the thickness of the fluid layer, $f = 2\Xi \sin\phi$ is the Coriolis parameter ($\Xi$ being the Earth's angular velocity), $g$ is the gravitational acceleration, and $\nu$ is the diffusion coefficient. The equations are defined over a spherical domain $\Omega = (\lambda, \phi)$, with longitude $\lambda$ and latitude $\phi$. 

As an initial condition, a zonal flow typical of a mid-latitude tropospheric jet is defined for the velocity component $u$ as a function of latitude $\phi$:
\[
u(\phi, t = 0) =
\begin{cases}
0, & \phi \leq \phi_0, \\
\frac{u_{\text{max}}}{n} \exp\left[\frac{1}{(\phi - \phi_0)(\phi - \phi_1)}\right], & \phi_0 < \phi < \phi_1, \\
0, & \phi \geq \phi_1,
\end{cases}
\]
where $u_{\text{max}}$ is the maximum zonal velocity, $\phi_0$ and $\phi_1$ represent the southern and northern boundaries of the jet in radians, and $n = \exp[-4/(\phi_1 - \phi_0)^2]$ normalizes $u_{\text{max}}$ at the midpoint of the jet. To induce barotropic instability, a localized Gaussian perturbation is added to the height field, expressed as:

\begin{align*}
h'(\lambda, \phi, t=0) =\hat{h} \cos(\phi) \exp\left[-\left(\frac{\lambda}{\alpha}\right)^2\right] \exp\left[-\left(\frac{\phi_2 - \phi}{\beta}\right)^2\right],    
\end{align*}
where $-\pi < \lambda < \pi$, and parameters $\hat{h}, \phi_2, \alpha$, and $\beta$ control the shape and location of the perturbation. The parameters $\alpha$ and $\beta$ are sampled from uniform distributions $\alpha \sim U[0.1, 0.5]$ and $\beta \sim U[0.03, 0.2]$.
We ran the solver from Dedalus~\cite{burns2020dedalus} on a $256\times256$ spherical grid, and the temporal dimension is discretized into 72 timestamps. We have a total of 200 trajectories by sampling $\alpha$ and $\beta$.

\begin{figure*}
\centering
\includegraphics[width=.99\linewidth]
{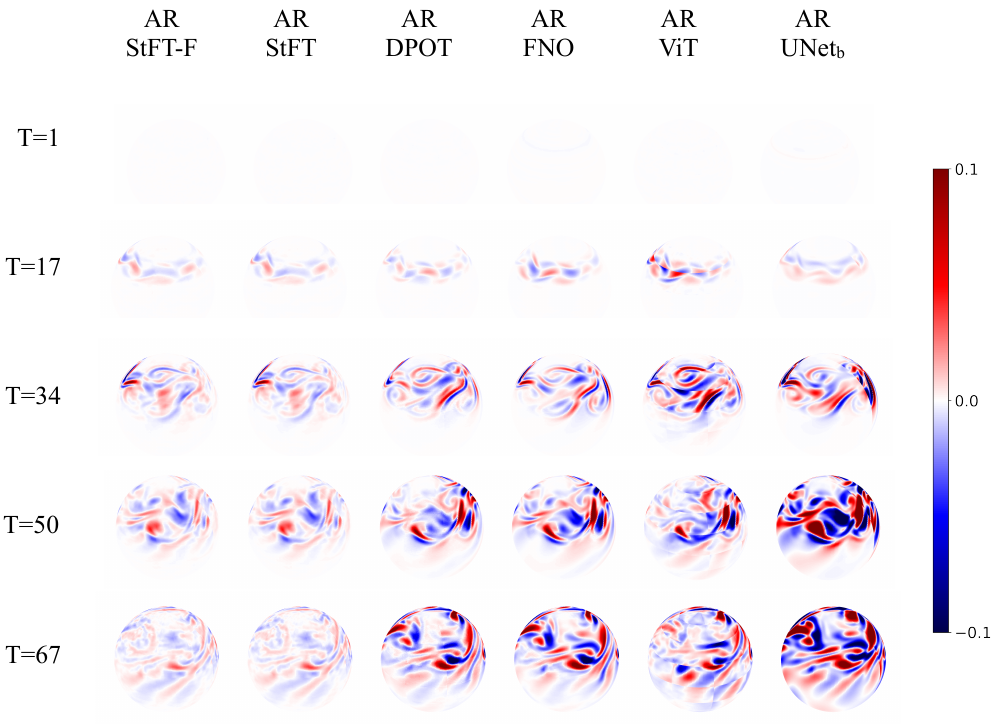}
\caption{\small Spherical shallow-water equations: pointwise error of the temporal evolution of velocity field predicted by all the autoregressive models: StFT-F, StFT, FNO, ViT and U-Net. The prediction error exhibits a temporal growth trend, with our model StFT and StFT-F consistently demonstrate lower residuals over the forecasting time horizon.} 
\label{fig:vis_sw}
\end{figure*}

\begin{figure*}[htp]
\centering
\includegraphics[width=\columnwidth]
{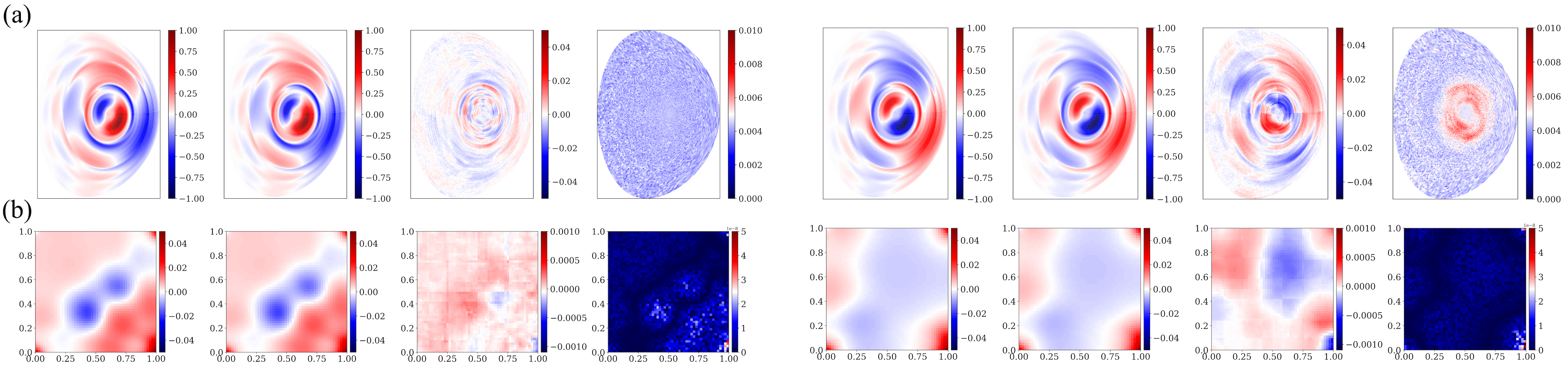}
\vspace{-.1in}
\caption{\small Additional evaluation of forecasting: ground truth, StFT-F prediction, residual, and uncertainty over time - shown for initial (left) and final (right) states. Variables include: (a) $\delta n_e$ in plasma MHD, (b) $p$ in Navier-Stokes.}
\label{fig:uq2}
\label{fig:}
\end{figure*}

\section{Ablation Study}\label{app2}


\subsection{The Hierarchical Structure and The Frequency Path}
To evaluate the effectiveness of the hierarchical structure and the frequency path in StFT, we conduct an ablation study of our model. First, we only keep one layer of \ours while removing the hierarchical structure. Second, we keep the hierarchical structure, and remove the frequency path in each hierarchical layer. 
\begin{table*}
\centering
\caption{Ablation study results. We run models with combinations of \ours blocks and the frequency path. $l_1$ is the coarsest level, and $l_3$ is the finest level. F stands for the frequency path in \ours blocks.}
\begin{tabular}{c|c|c}
\toprule
\textbf{Dataset} & \textbf{Setting(s)} & \textbf{AR-\ours}\\
\hline
\multirow{4}{*}{Plasma MHD}
& $l_1 + \mathcal{F}$ & 0.0805
 \\
& $l_2 + \mathcal{F}$ &0.105  \\
& $l_1+l_2$ & 0.0404\\
& $l_1+l_2+\mathcal{F}$ &\textbf{0.0307}\\
\hline
\multirow{5}{*}{Shallow-Water} 
& $l_1+\mathcal{F}$ & 2.5729 \\
& $l_2+\mathcal{F}$& 0.101 \\
& $l_3+\mathcal{F}$  &0.0975\\
& $l_1+l_2+l_3$ &0.0956\\
& $l_1+l_2+l_3+\mathcal{F}$ &\textbf{0.0625}\\
\bottomrule
\end{tabular}
\label{tb:ablation0}
\end{table*}

Table~\ref{tb:ablation0} shows the $L_2$ relative errors averaged over all the variables. 
Note that $l_1$ is the coarsest level, and $l_3$ is the finest level. $\mathcal{F}$ stands for the frequency path in \ours blocks. We find that multi-scale $+\mathcal{F}$" outperforms both ablation cases.
These results demonstrate the effectiveness of the hierarchical composition of \ours blocks and the frequency path in \ours block. With the hierarchical composition, for Plasma MHD, the error drops to $0.0307$ from $0.0805$, and for shallow-water equations (SWE), the error drops to $0.0625$ from $0.0975$. 
We observe that the fine-level layer setting in the SWE achieves the best performance among the single layer results, and the multi-layer settings further decrease the prediction error.
The frequency path in \ours also plays a crucial role, where in Plasma MHD, the error drops to $0.0307$ from $0.0404$, and in the SWE dataset, the error drops to $0.0625$ from $0.0956$. 


\subsection{The Overlapping Tokenizer}

We conduct an ablation study on the overlapping tokenizer design, evaluating both predictive performance and computational cost, as reported in Tables~\ref{tb:ot_err} and~\ref{tb:ot_comp}.We evaluate the computational cost including the inference FLOPS per sample, training/inference time per batch and peak memory usage. We used a fixed batch size of 20 for all models. The results are summarized in the following two tables. StFT-O stands for StFT with the overlapping tokenizer and StFT-NO for the opposite. All experiments were performed on an NVIDIA A100 GPU. Incorporating the overlapping tokenizer leads to substantial improvements in accuracy, in plasma MHD and shallow-water datasets, with errors reductions of $68\%$ and $10\%$ respectively. Notably, this enhancement comes with negligible impact on computational complexity, inference time and peak memory usage.

\begin{table}[htbp]
\centering
\caption{Effect of the overlapping tokenizer on the prediction error of plasma MHD and shallow-water datasets.}
\begin{tabular}{c|c|c}
\toprule
\textbf{Method} & \textbf{Plasma MHD} & \textbf{Shallow-Water} \\
\midrule
With overlapping tokenizer & \textbf{0.0307} & \textbf{0.0625} \\
Without overlapping tokenizer & 0.0986 & 0.0700 \\
\bottomrule
\end{tabular}
\label{tb:ot_err}
\end{table}

\begin{table}[htbp]
\centering
\setlength\tabcolsep{3.5pt}
\caption{Computational complexity comparison of StFT with and without overlapping tokenizer.}
\begin{tabular}{c|c|c|c|c|c}
\toprule
\textbf{Dataset} & \textbf{Method} & \textbf{GFLOPs} & \textbf{Training / Iter (s)} & \textbf{Inference / Iter (s)} & \textbf{Peak Memory (GB)} \\
\midrule
MHD & StFT-O  & 0.704 & 0.181 & 0.0455 & 9.49 \\
MHD & StFT-NO & 0.704 & 0.157 & 0.0454 & 9.49\\
\midrule
SWE & StFT-O  & 4.30 & 0.189 & 0.0319 & 12.2 \\
SWE & StFT-NO & 4.30 & 0.189 & 0.0319 & 12.2\\
\bottomrule
\end{tabular}
\label{tb:ot_comp}
\end{table}

\subsection{Convergence on the Number of Scales}

We evaluate the performance of StFT on the number of scales for the plasma MHD dataset. Specifically, we run \ours with one scale (patch size of $128$), three scales (patch sizes of 128, 64, and 48), and four scales (patch sizes of 128, 64, 48, and 32). As shown in Table~\ref{tb: scales}, it is observed that increasing the number of scales leads to a reduction in prediction error to a certain point, with the two-layer configuration achieving the lowest error ($0.0307$). Beyond this, the inclusion of additional scales results in a slight increase in error, indicating a convergence in model performance, and further increasing the number of scales may not provide performance gain.

\begin{table}[htbp]
\centering
\caption{Effect of the number of scales on L2 relative error of the overall prediction on the plasma MHD.}
\begin{tabular}{cc}
\toprule
\textbf{Number of Scales} & \textbf{L2 Relative Error} \\
\midrule
1 & 0.0805 \\
2 & 0.0307 \\
3 & 0.0385 \\
4 & 0.0391 \\
\bottomrule
\end{tabular}
\label{tb: scales}
\end{table}

\section{Computational and Space Complexity}\label{computation}

\begin{figure*}
\centering
\includegraphics[width=.99\linewidth]
{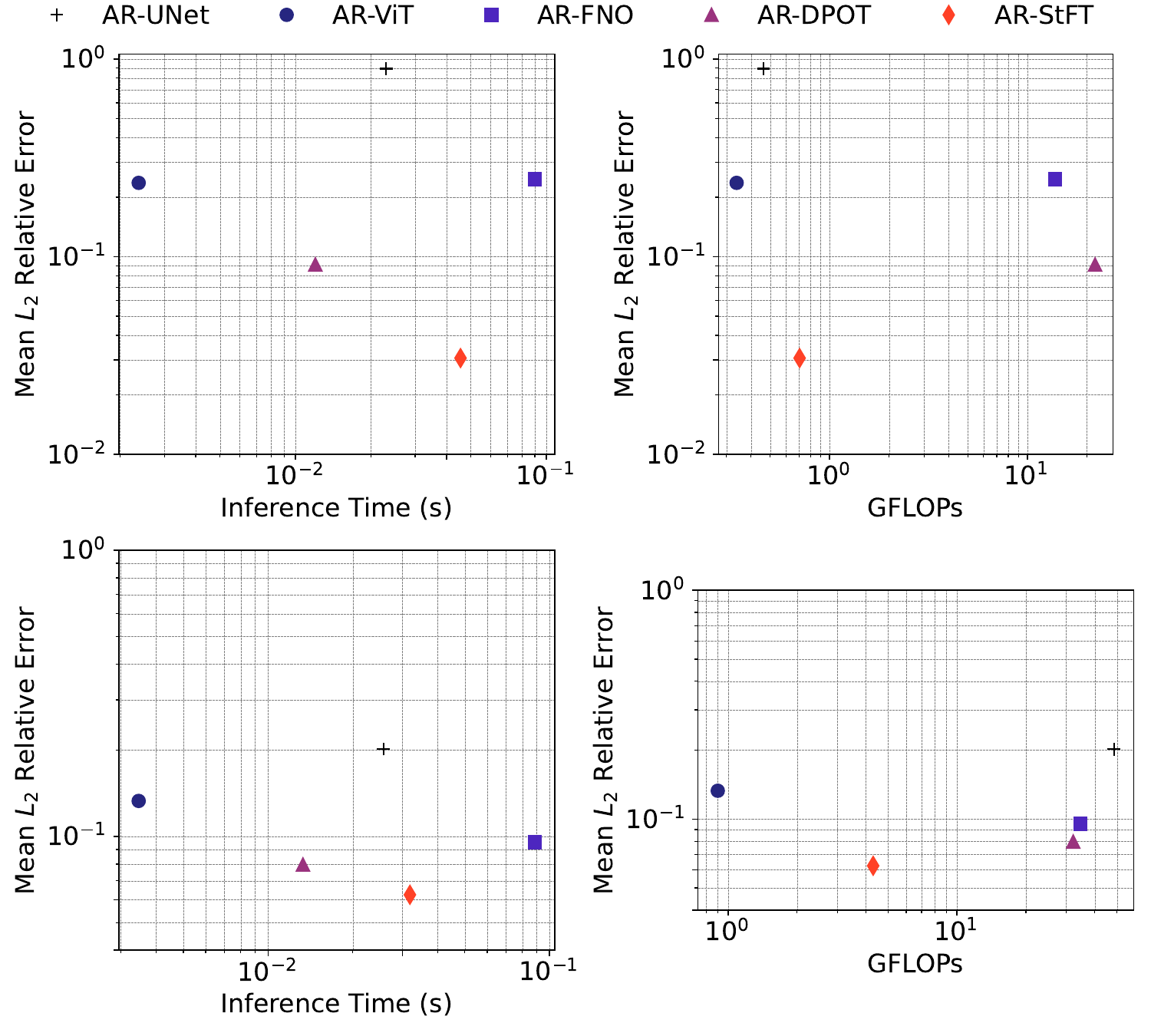}
\caption{\small Comparison of models in terms of mean L2 relative error versus inference time (left) and FLOPs (right) for plasma MHD (top) and shallow-water (bottom).} 
\label{fig:flops_inference}
\end{figure*}

We compare \ours with other baselines regarding the inference FLOPS per sample, training time per batch, inference time per batch, and peak memory usage (during training). In plasma MHD and shallow-water datasets, compared to FNO, StFT is $20$x smaller, and $8$x smaller in computation (FLOPS) respectively. While achieving the highest prediction accuracy, StFT has the same order of magnitude of FLOPs as UNet and ViT in plasma MHD. In the shallow-water equation (SWE), however UNet is much more computationally expensive - $11$x more than StFT – while its relative error is $3$x larger than StFT. StFT is also efficient in both training and inference time. For plasma MHD, StFT achieves a $30\%$ reduction in training time and $50\%$ reduction in inference time compared to FNO. For SWE, StFT is $40\%$ faster in training and $60\%$ faster than FNO in inference.  Although StFT incorporates dual paths operating in the frequency domain and the spatio-temporal domain, its peak memory usage remains comparable to that of FNO in the plasma MHD, and is reduced by $56\%$ compared to FNO in SWE, and $26\%$ less than UNet in SWE. Figure \ref{fig:flops_inference} compares the inference time and FLOPs versus mean L2 relative error for all models.

\begin{table}[htbp]
\centering
\caption{Comparison of methods on plasma MHD and shallow-water datasets regarding the computational and space complexity.}
\begin{tabular}{l|c|c|c|c|c}
\toprule
\textbf{Dataset} & \textbf{Method} & \textbf{GFLOPs} & \textbf{Training / Iter (s)} & \textbf{Inference / Iter (s)} & \textbf{Peak Memory (GB)} \\
\midrule
MHD & StFT  & 0.704  & 0.181 & 0.0455 & 9.49 \\
MHD& DPOT   & 1.10  & 0.0676 & 0.0120 & 3.20 \\
MHD& FNO   & 13.8  & 0.262 & 0.0900 & 10.1 \\
MHD& UNET  & 0.462  & 0.0263 & 0.00230 & 1.38 \\
MHD& ViT   & 0.338  & 0.0171 & 0.00237 & 1.65 \\
\midrule
SWE & StFT  & 4.30  & 0.189 & 0.0319 & 12.2 \\
SWE & DPOT  & 1.61  & 0.0427 & 0.0133 & 8.43 \\
SWE & FNO   & 34.6  & 0.325 & 0.0886 & 30.2 \\
SWE & UNET  & 48.5  & 0.107 & 0.0257 & 16.5 \\
SWE & ViT   & 0.90  & 0.0130 & 0.00347 & 7.45 \\
\bottomrule
\end{tabular}
\end{table}

\section{Signficance of UQ on StFT-F}\label{sec:UQ}

For the Plasma MHD dataset, we measure the average coverage for the first $20$ autoregressively predicted snapshots (each row represents one physical variable), as shown in Table~\ref{tb:ci-plasma}. In the last row, we report the average coverage across all physical variables. For the $90\%$ confidence interval, average coverage is very close to the ideal value of $90\%$, demonstrating that StFT-F is well-calibrated around the $90\%$ confidence interval. For the $95\%$ confidence interval, the average coverage is under-confident by $2.5\%$, suggesting the intervals may be slightly narrow. For the shallow-water dataset, we provide the results in Table \ref{tb:ci-sw}, where we measure the average coverage for the first $10$ predicted snapshots. For the $90\%$ confidence interval, average coverage is also very close to the ideal value of $90\%$, which shows that StFT-F is also well-calibrated for the shallow-water dataset.

\begin{table}[htbp]
\centering
\caption{Confidence interval coverage on the plasma MHD dataset.}
\begin{tabular}{ccc}
\toprule
\textbf{Variable} & \textbf{CI: 90\%} & \textbf{CI: 95\%} \\
\midrule
$\delta\phi$    & 0.906 & 0.937 \\
$\delta A_{\parallel}$ & 0.977 & 0.988 \\
$\delta B_{\parallel}$ & 0.900 & 0.930 \\
$\delta n_e$    & 0.933 & 0.962 \\
$\delta n_i$    & 0.867 & 0.910 \\
$\delta u_e$    & 0.781 & 0.823 \\
\midrule
\textbf{Average Coverage} & \textbf{89.4\%} & \textbf{92.5\%} \\
\bottomrule
\end{tabular}
\label{tb:ci-plasma}
\end{table}

\begin{table}[htbp]
\centering
\caption{Confidence interval coverage on the shallow-water dataset.}
\begin{tabular}{ccc}
\toprule
\textbf{Variable} & \textbf{CI: 90\%} & \textbf{CI: 95\%} \\
\midrule
$\mathbf{V}$ & 0.895 & 0.980 \\
\bottomrule
\end{tabular}
\label{tb:ci-sw}

\end{table}

\begin{table}[h!]
\centering
\renewcommand{\arraystretch}{1.2}
\caption{Training/validation/test data splits and the training budget for all models measured on an A100 GPU.}
\begin{tabular}{l|c|c|c}
\hline
\textbf{Dataset} & \textbf{Total} & \textbf{Split (Train / Val / Test)} & \textbf{Training Budget} \\
\hline
Plasma MHD & 1 traj. \newline (1,221 snapshots) & 927 / 50 / 244  & 24h \\
\hline
Navier-Stokes & 100 traj. \newline (101 snaps/traj) & 80 / 10 / 10 & 24h \\
\hline
Shallow-water & 200 traj. \newline (72 snaps/traj) & 160 / 20 / 20 & 48h \\
\hline
\end{tabular}
\label{tb:summary_training}
\end{table}

\section{Experimental Details}\label{app3}
\textbf{Training/validation/test data sets. } For the plasma MHD data, we split the trajectory of $1221$ snapshots into a training set (the first $927$ snapshots), a validation set (the middle $50$ snapshots), and a test set (the last $244$ snapshots). For the Navier-Stokes dataset, we have a total of $100$ trajectories ($101$ snapshots for each trajectory), and split them into $80$ trajectories for training, $10$ for validation, and $10$ for testing. For the shallow-water dataset, we have a total of $200$ trajectories ($72$ snapshots for each trajectory), and split them into $160$ trajectories for training, $20$ for validation, and $20$ for testing. To ensure a fair comparison, we impose a fixed training budget across all models. Specifically, a $48$-hour limit mesaured on one A100 GPU was set for the shallow-water equation dataset, while a $24$-hour limit is applied to both the plasma MHD and Navier-Stokes equation datasets. Table ~\ref{tb:summary_training} summarizes the datasets and the training budget for all models.


\textbf{Generative residual correction block. 
} We follow a two-step training protocol in training \ours-F: first, we train \ours thoroughly with the training budget, and then we train the generative residual correction block for another 200 epochs. We employ a rectified flow to learn distributions of the residuals given the prediction of \ours and the history snapshots. We implement a similar structure to the Diffusion Transformer (DiT) as the backbone model~\citep{peebles2023scalable}. In each DiT block, we apply adaptive layer normalization before a self-attention layer and an MLP layer. We use adaLN-Zero for time conditioning. For the history snapshots $\tilde{u}_{t}$ and the prediction $\mathcal{F}_{\theta_d}(\tilde{u}_{t})$ of \ours, these conditions are incorporated as extra input tokens.


\textbf{Hyperparameters. }For \ours on the plasma dynamics, 3D FFT is used to encode the spatio-temporal inputs in the frequency path. We use the patch size of $128$ for the the first \ours block and and $64$ for the second \ours block. The overlapping size is set to $1$. The hidden dimension is set to $128$. The depth for each \ours block is set to $6$. We keep the lowest $8$ frequencies for each spatial dimension. For the rectified flow block, the depth is set to $8$, and the hidden dimension is set to $128$. For the Navier-Stokes equation, \ours uses a patch size of $25$ for the coarse block (the first block), $13$ for the middle block, and $8$ for the last block. The overlapping size is set to $0$, and the frequency path is not used. For each block in the hierarchical structure, the depth is set to 8, and we use a hidden dimension of 512. In the rectified flow block, we use a depth of $4$ and set the hidden dimension to $128$. For the shallow-water equation, three levels of \ours blocks are employed, and their patch sizes are set to $128$, $64$, and $32$ respectively. For each block, the depth is set to $6$, and the hidden dimension is set to $512$. We use the 2D FFT to encode the spatio-temporal inputs, and the lowest $8$ frequencies are kept for each spatial dimension. The overlapping size is set to $1$. For the rectified flow model, we use a depth of $8$ and a hidden dimension of $128$.

\begin{table}[h]
\centering
\caption{Hyperparameter search range for each method.}
\begin{tabular}{c|l}
\hline
\textbf{Method} & \textbf{Hyperparameter Search } \\
\hline
\multirow{4}{*}{\textbf{DPOT}} 
    & Hidden dimension: $[256, 512]$ \\
    & Patch size: $[8, 16, 32]$ \\
    & Depth: 6\\
    & Heads: 4  \\
\hline
\multirow{3}{*}{\textbf{AR-FNO}} 
    & Modes: $[16, 20, 24]$ \\
    & Layers: $[4, 5]$ \\
    & Hidden dimension: 256 \\
\hline
\multirow{2}{*}{\textbf{AR-ViT}} 
    & Hidden dimension: $[256, 512]$ \\
    & Patch size: $[16, 32, 64]$ \\
\hline
\multirow{1}{*}{\textbf{AR-UNet}} 
    & Bottleneck hidden dimension: $[64, 128, 256, 512]$ \\
\hline
\end{tabular}
\label{tb:grid_search}
\end{table}

 \textbf{Baselines. } For all the baselines, we run all models with the same training time budget, as detailed in Table ~\ref{tb:summary_training}. For DPOT, we vary the hidden dimension from $[256,512]$, and the patch size from $[8,16,32]$. We set the depth as 6, and the number of heads as 4, which is the default setting from the author's implementation. For AR-FNO, the number of modes are selected through a search in $[16,20,24]$, the number of layers are searched in $[4,5]$, and the hidden dimension is set to 256. For AR-ViT, we vary the hidden dimension from $[256,512]$, and the patch size from $[16,32,64]$. For AR-UNet, the hidden dimension of bottleneck embeddings are searched in $[64,128,256,512]$. Table ~\ref{tb:grid_search} summarizes the hyperparameter choices for each method.

\section{More Visualization Results}\label{app4}
Figure~\ref{fig:vis_plasma} illustrates the ground truth and predicted temporal evolution of normalized perturbed parallel vector potential $\delta A_{\parallel}$ and perturbed electron density $\delta n_e$ in plasam MHD using AR-\ours, AR-\ours-F, AR-DPOT, AR-FNO, AR-ViT and AR-UNet methods. StFT and StFT-F perform stable across the forecasting time horizon.  Figure~\ref{fig:vis_ns} and~\ref{fig:vis_sw} show the pointwise error of all the models compared to the ground truth data in the Navier-Stokes and shallow-water equations, where StFT and StFT-F demonstrate lower residuals compared to other baseline models.

\begin{figure*}[htp]
\centering
\includegraphics[width=0.8\columnwidth]
{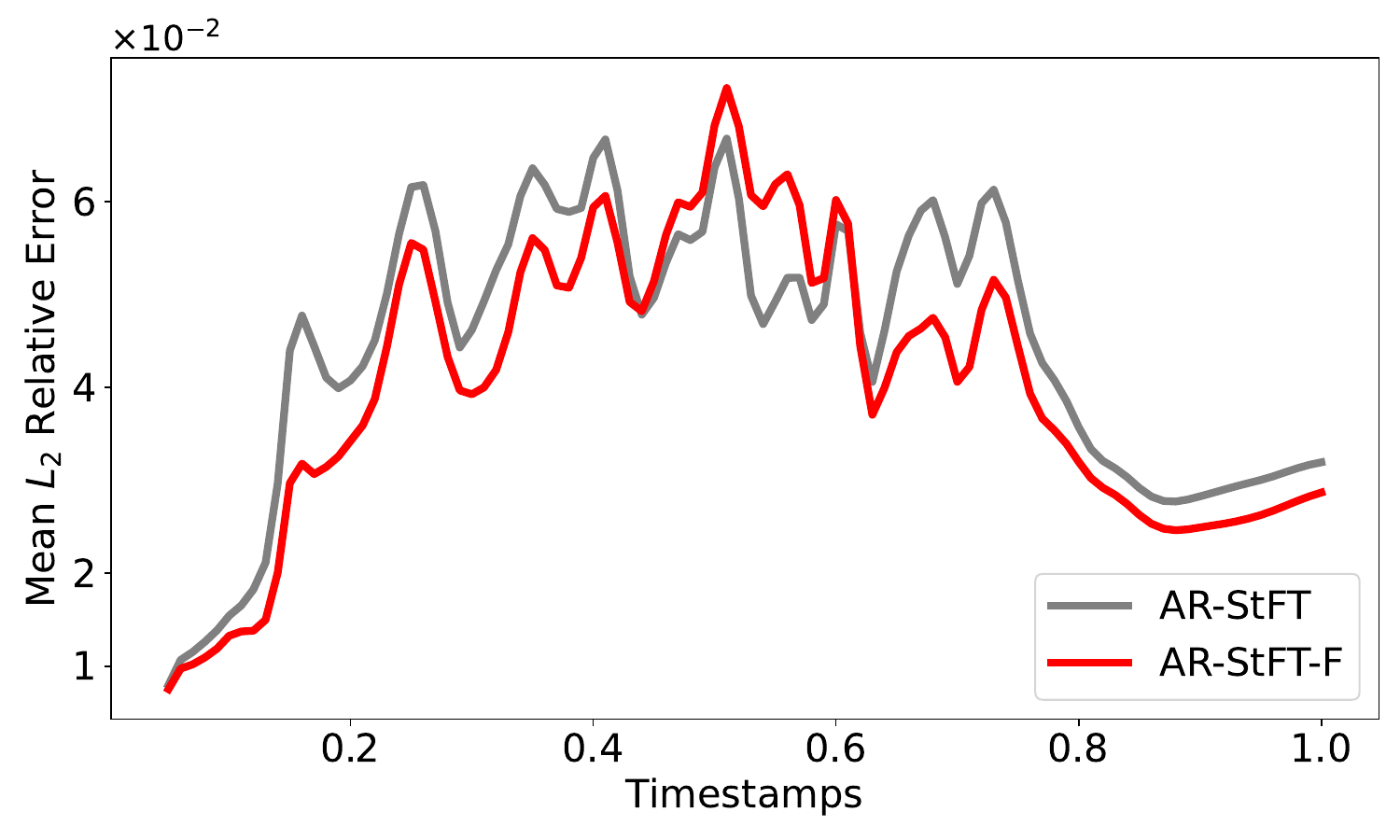}
\vspace{-.1in}
\caption{\small Results of comparing StFT and StFT-F for the autoregressive prediction in $L_2$ relative error across different timestamps for velocity $u$ in Navier-Stokes equation. The shaded region represents the standard deviation distribution of the relative error of StFT-F. However, the uncertainty is negligible that it is not visually discernible. StFT-F demonstrates better performance in the latter stages of the forecasting time horizon compared to StFT.}
\label{fig:ns_compare_stft}
\end{figure*}

In addition, we compare \ours with \ours-F for the autoregressive prediction across different timestamps for velocity $u$ in Navier-Stokes equation, shown in Figure ~\ref{fig:ns_compare_stft}. A slightly larger error is observed 
between $t=0.4$ and $0.6$, which can be attributed to the training objective of StFT. As StFT is optimized using a pointwise loss function, it is encouraged to produce predictions that closely match the most likely outcome. In contrast, StFT-F is designed to learn the full distribution of the target, potentially introducing higher variance in its predictions. This distributional modeling, while beneficial for uncertainty quantification, may result in marginally increased errors as it captures characteristics beyond the mean behavior.

\end{document}